\documentclass{article}

\usepackage{PRIMEarxiv}

\usepackage[utf8]{inputenc} 
\usepackage[T1]{fontenc}    
\usepackage{hyperref}       
\usepackage{url}            
\usepackage{booktabs}       
\usepackage{amsfonts}       
\usepackage{nicefrac}       
\usepackage{microtype}      
\usepackage{lipsum}
\usepackage{fancyhdr}       
\usepackage{graphicx}       
\graphicspath{{media/}}     

\usepackage{amsmath}
\usepackage{amssymb}
\usepackage{subcaption}
\usepackage{overpic}
\usepackage{color}
\usepackage[dvipsnames]{xcolor}
\usepackage{pdfpages}
\usepackage{comment}
\usepackage{multirow}
\usepackage{multicol}
\usepackage{threeparttable}
\usepackage{framed}
\usepackage{float}
\usepackage{bbding}
\usepackage{pifont}
\usepackage{booktabs}
\usepackage{makecell}
\usepackage{titlesec}
\usepackage{authblk}
\usepackage[ruled,linesnumbered]{algorithm2e}
\usepackage{algorithmic}
\usepackage[capitalize]{cleveref}

\newcommand\ie{\textit{i.e.}}
\newcommand\etal{\textit{et al.}}

\makeatletter
\newcommand*\bigcdot{\mathpalette\bigcdot@{.5}}
\newcommand*\bigcdot@[2]{\mathbin{\vcenter{\hbox{\scalebox{#2}{$\m@th#1\bullet$}}}}}
\makeatother

\title{CORP: A Multi-Modal Dataset for Campus-Oriented Roadside Perception Tasks}

\author[1]{Beibei Wang$^{\star}$}
\author[2]{Zijian Yu$^{\star}$}
\author[1]{Lu Zhang}
\author[1]{Jingjing Huang}
\author[2]{Yao Li}
\author[2]{Haojie Ren}
\author[2]{Yuxuan Xiao}
\author[2]{Yuru Peng}
\author[2]{Jianmin Ji$^{1,}$}
\author[2]{Yu Zhang$^{1,}$}
\author[2]{Yanyong Zhang$^{1,}$}

\affil[$\star$]{equal contribution}
\affil[1]{Institute of Artificial Intelligence, Hefei Comprehensive National Science Center}
\affil[2]{University of Science and Technology of China}

\begin{document}
\maketitle

\begin{abstract}
Many roadside perception datasets have been proposed to facilitate the development of autonomous driving and intelligent traffic technologies. However, it has been observed that the majority of their concentrates is on urban arterial roads, inadvertently overlooking residential areas such as parks and campuses that exhibit entirely distinct characteristics. In light of this gap, we propose CORP, which stands as the first public benchmark dataset tailored for multi-modal roadside perception tasks under campus scenarios. Collected in a university campus, CORP consists of over 205k images plus 102k point clouds captured from 18 cameras and 9 LiDAR sensors. These sensors with different configurations are mounted on roadside utility poles to provide diverse viewpoints within the campus region. The annotations of CORP encompass multi-dimensional information beyond 2D and 3D bounding boxes, providing extra support for 3D seamless tracking and instance segmentation with unique IDs and pixel masks for identifying targets, to enhance the understanding of objects and their behaviors distributed across the campus premises. Unlike other roadside datasets about urban traffic, CORP extends the spectrum to highlight the challenges for multi-modal perception in campuses and other residential areas.
\end{abstract}


\section{Introduction}\label{sec:intro}

Roadside perception (RP) technology is a key component in the intelligent transportation and autonomous driving domains, enabling features such as traffic flow management\cite{roadside_monitor}, decision support for autonomous vehicles\cite{vips}, and campus surveillance and security\cite{a_day_on_campus}, where the pursuit of reliable perception methods and high-quality datasets remain enduring topics.

To propel advancements in roadside perception technology, numerous roadside datasets have been created with a primary focus on high-density traffic flows along urban arterial roads\cite{yu2022dairv2x,rope3d_2022_CVPR,a9i_2023}, intersections\cite{a9i_2023,isp300_2022} and highway\cite{a9_2022}. However, within university campuses and similar residential areas such as public parks, we found the environment faced in roadside perception tasks exhibits significant differences:

\begin{itemize}
\item Compared with the dominant flow of motorized vehicles in urban traffic, a higher density of active pedestrians and cyclists appears in campus scenarios.
\item Unlike the linearly structured urban arterial roads, residential campuses present distinct physical layouts characterized by tightly packed clusters of buildings, lush vegetative areas, narrower paths with more frequent corners, and vehicle-restricted zones that complicate navigation.
\item The lack of traffic ordinances and control systems introduces considerable intricacies into the analysis of user trajectories and behavioral patterns.
\end{itemize}

\begin{figure}[tb]
\centering

    \begin{minipage}[b]{.32\columnwidth}
        \centering
        \includegraphics[width=1.0\columnwidth, height=0.56\columnwidth]{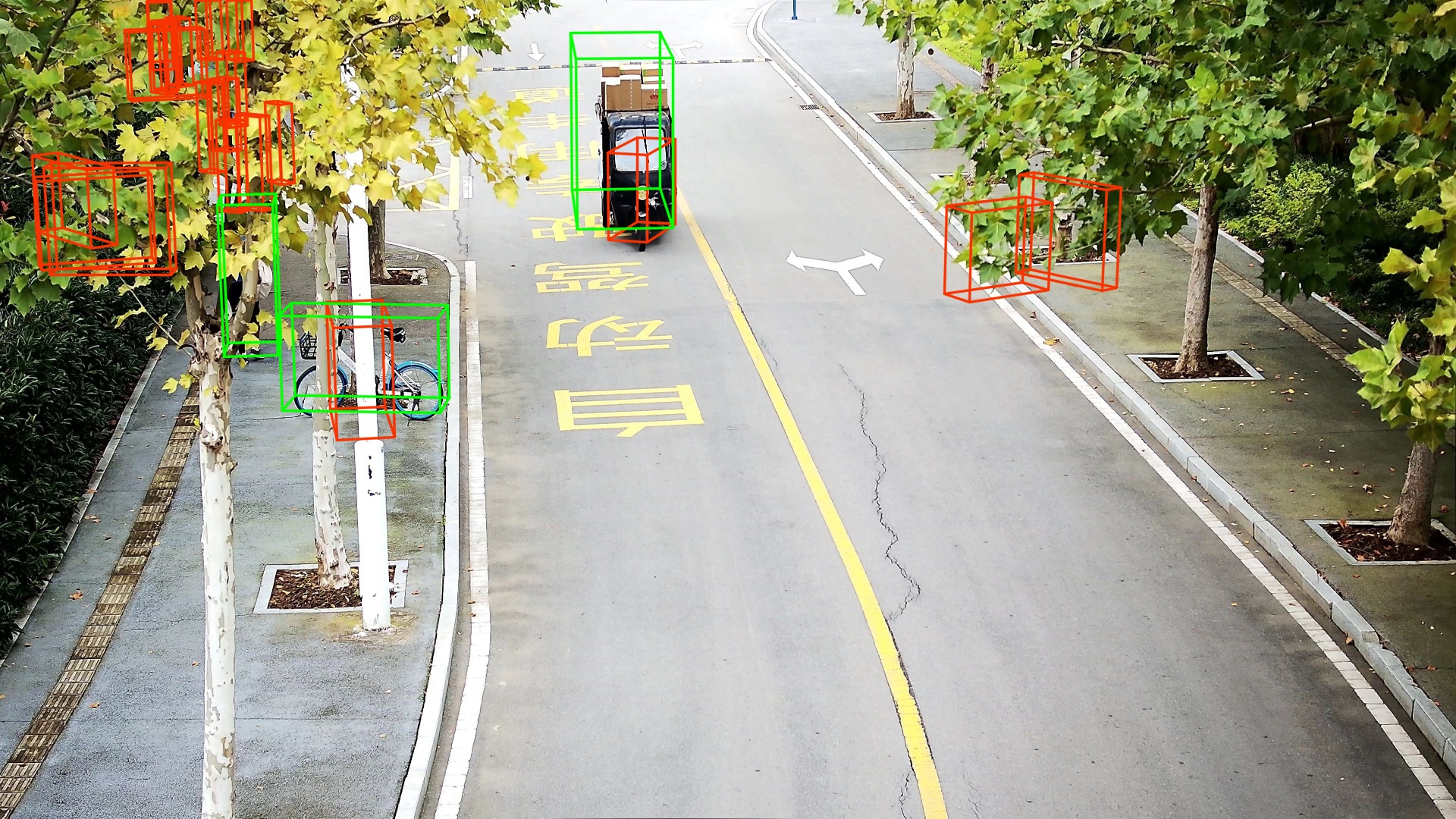} \\
        \centering
        \includegraphics[width=1.0\columnwidth, height=0.56\columnwidth]{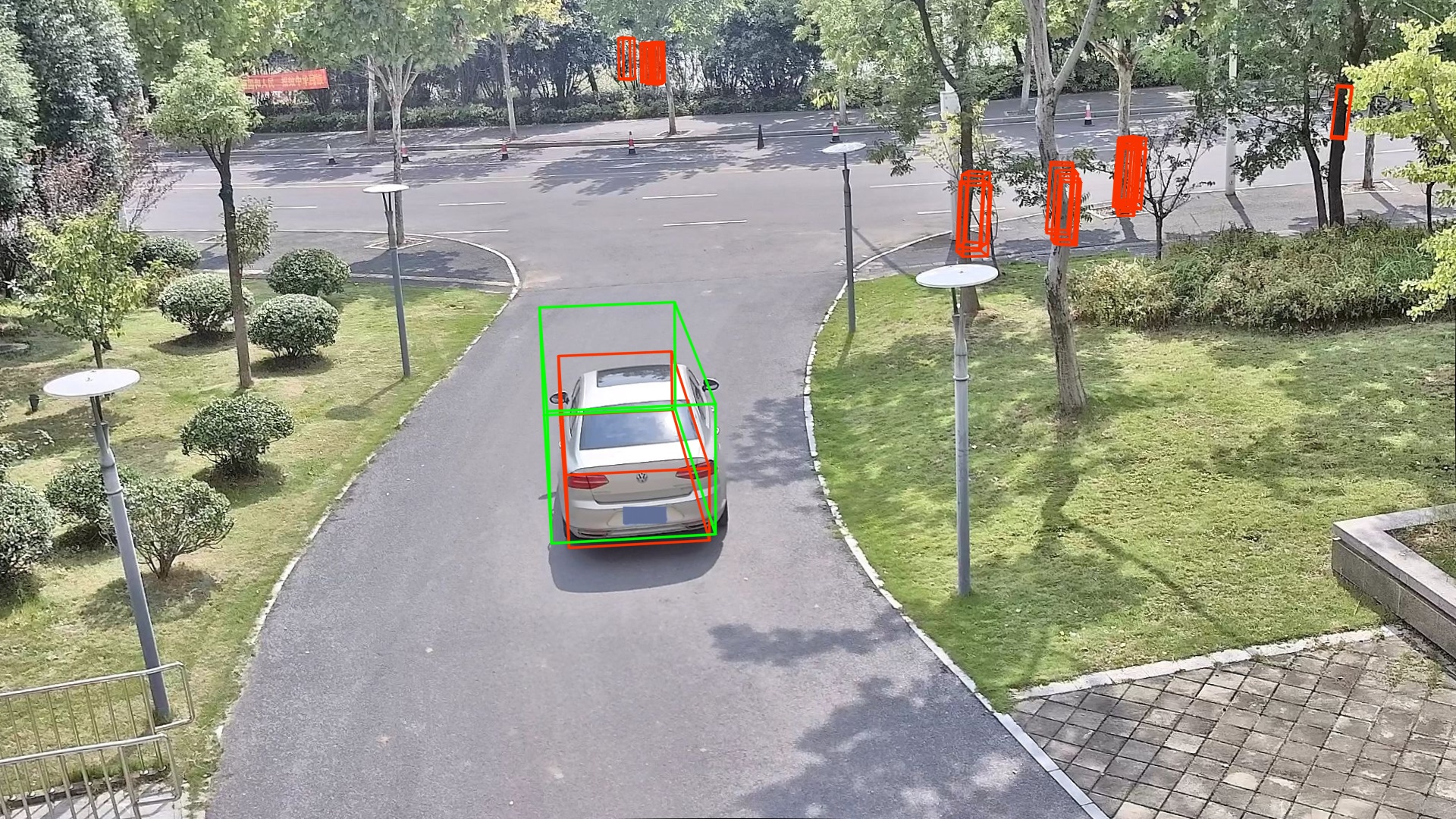} 
        \subcaption{BEVDepth Pred. vs GT}
    \end{minipage}\hspace{0pt}
    \begin{minipage}[b]{.32\columnwidth}
        \centering
        \includegraphics[width=1.0\columnwidth, height=0.56\columnwidth]{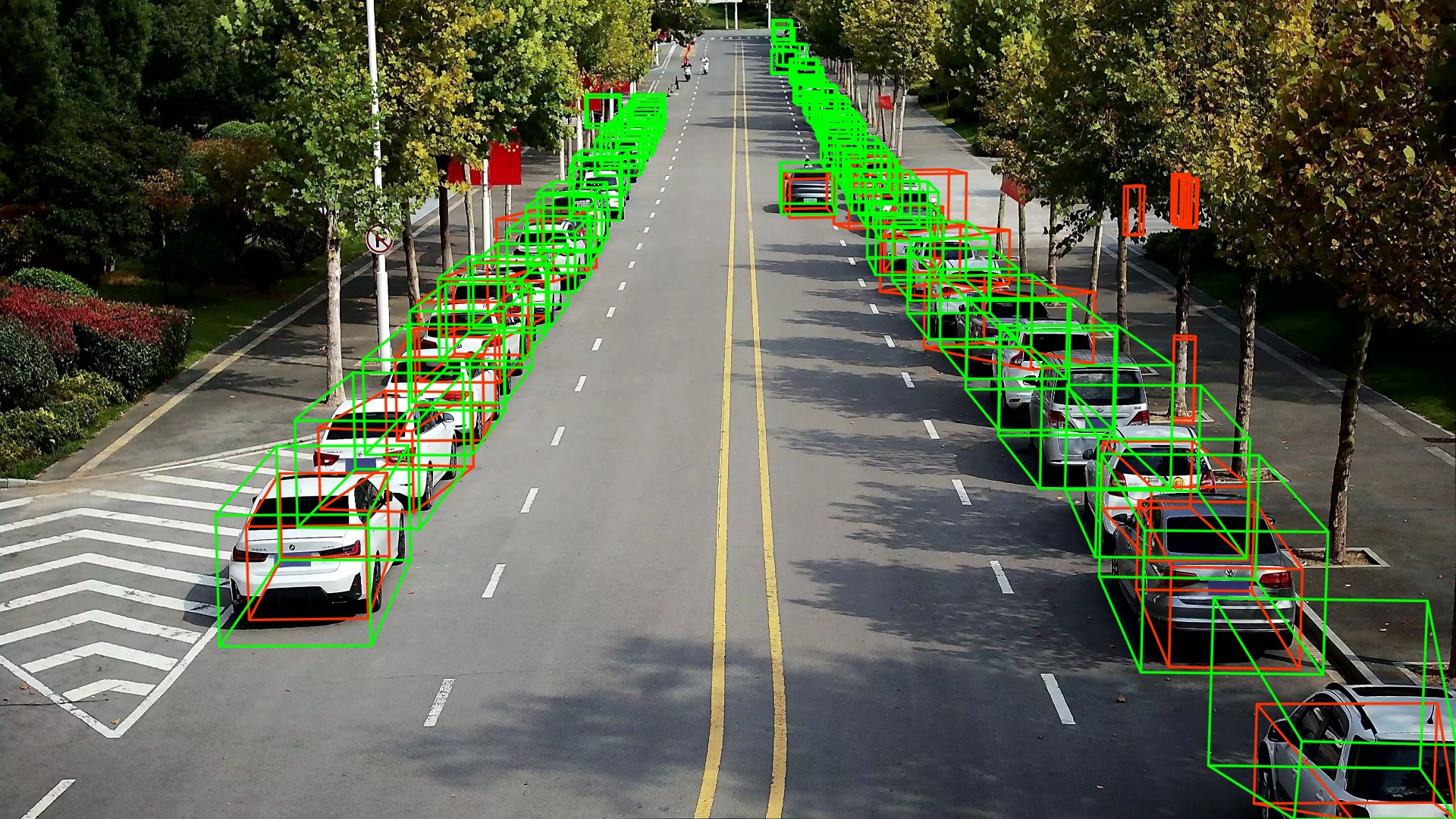} \\
        \centering
        \includegraphics[width=1.0\columnwidth, height=0.56\columnwidth]{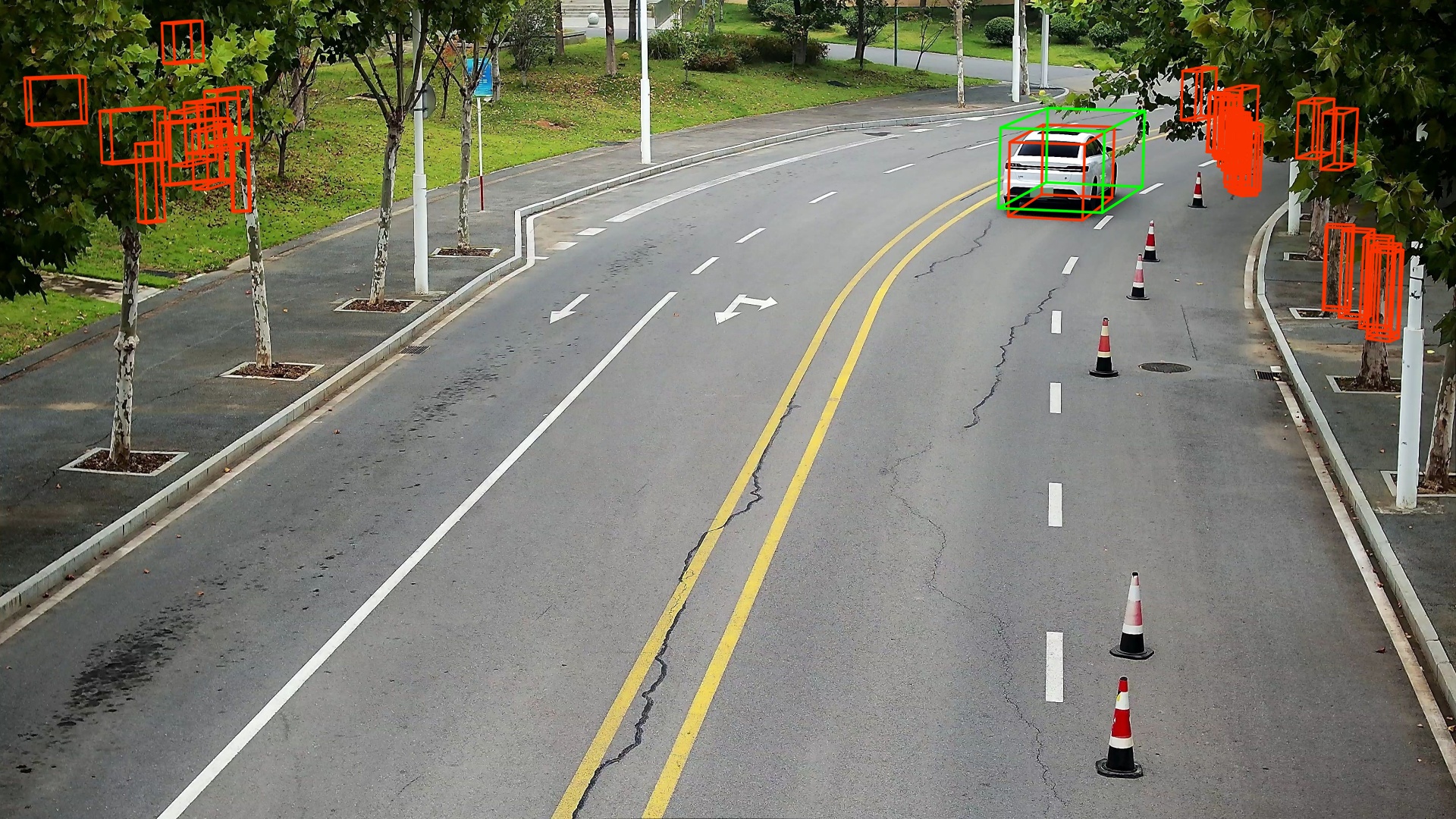} 
        \subcaption{BEVHeight Pred. vs GT}
    \end{minipage}\hspace{0pt}
    \begin{minipage}[b]{.32\columnwidth}
        \centering
        \includegraphics[width=1.0\columnwidth, height=0.56\columnwidth]{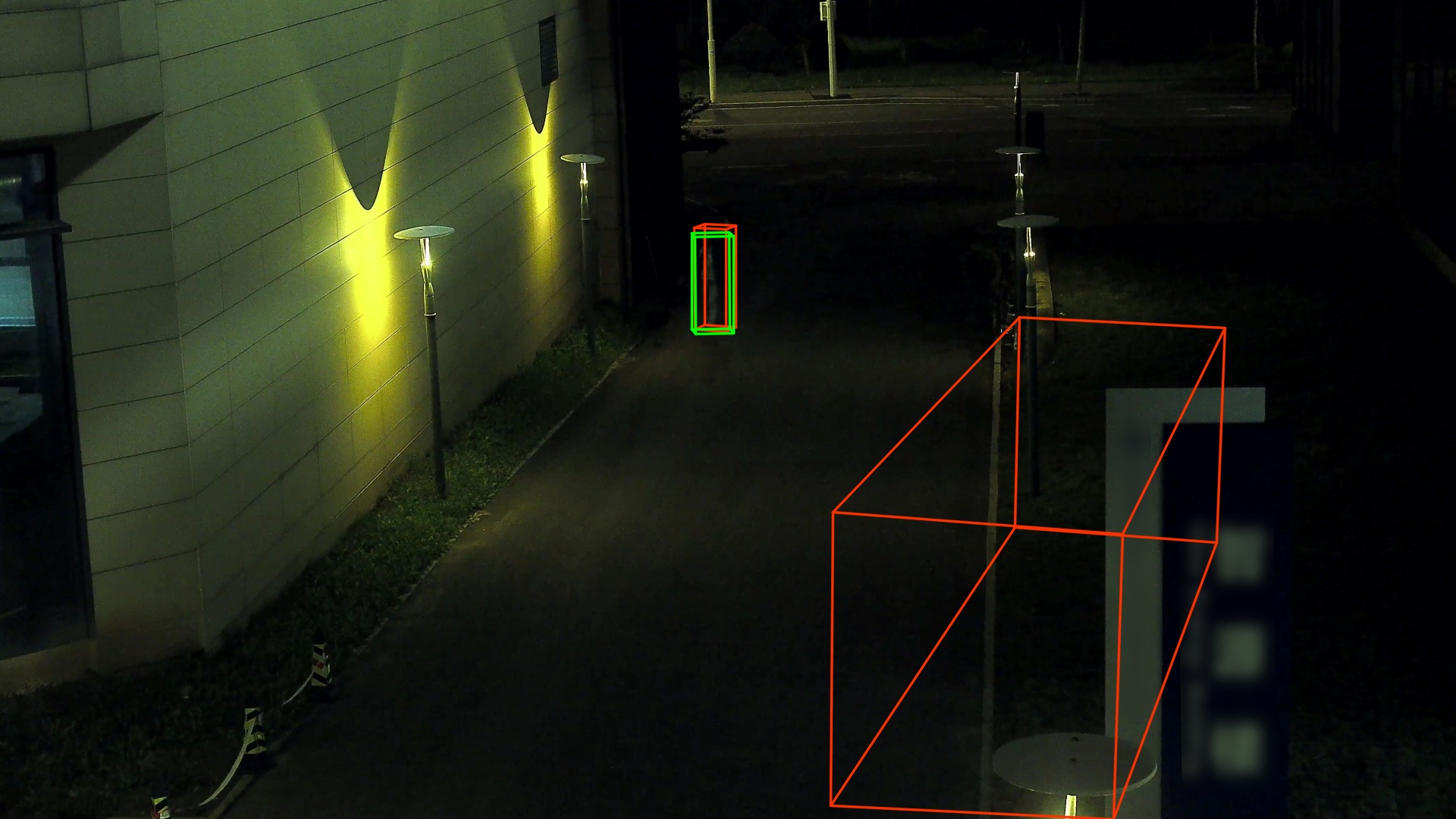} \\
        \centering
        \includegraphics[width=1.0\columnwidth, height=0.56\columnwidth]{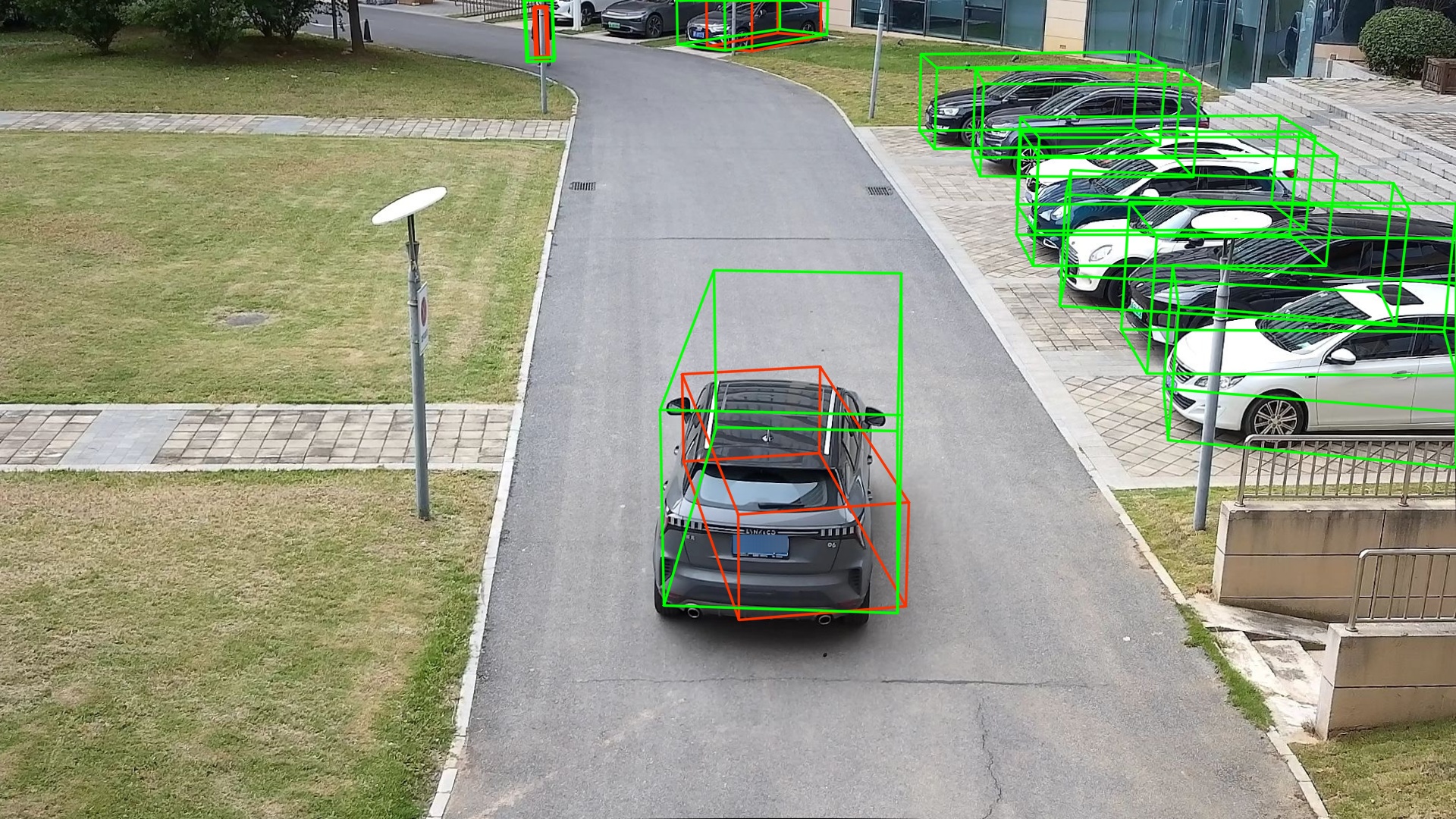} \\
        \subcaption{BEVFormer Pred. vs GT}
    \end{minipage}
\caption{In the domain of 3D object detection, BEVDepth\cite{li2022bevdepth}, BEVFormer\cite{li2022bevformer} and BEVHeight\cite{yang2023bevheight} stand out, which were tested in a real-world campus setting. The green bounding boxes represent the ground truth annotations (GT), while the red boxes denote the predicted results (Pred.). Due to the relatively sparse representation of natural environment elements during the training phase using DAIR-V2X-I\cite{yu2022dairv2x} dataset, these algorithms struggle to efficiently discern target objects amidst lush green vegetation in practical applications.}
\label{fig:dair_model_test}
\end{figure}

On a university campus, we have tested some open-sourced state-of-the-art algorithms in the roadside perception field such as BEVHeight\cite{yang2023bevheight}, BEVDepth\cite{li2022bevdepth} and BEVFormer\cite{li2022bevformer} with DAIR-V2X-I\cite{yu2022dairv2x} and corresponding pre-trained models. When applied to campus roads characterized by complex living scenes and natural elements, these algorithms demonstrated unsatisfactory performance, failing to effectively distinguish between various objects and backgrounds, as illustrated in \cref{fig:dair_model_test}. This phenomenon indicates that, despite their outstanding performance on urban traffic datasets like DAIR-V2X\cite{yu2022dairv2x} and Rope3D\cite{rope3d_2022_CVPR}, these algorithms still need improvement in terms of generalization capability and robustness when faced with more diverse environments.

\begin{figure}[tb]
\begin{overpic}[width=1.0\columnwidth]{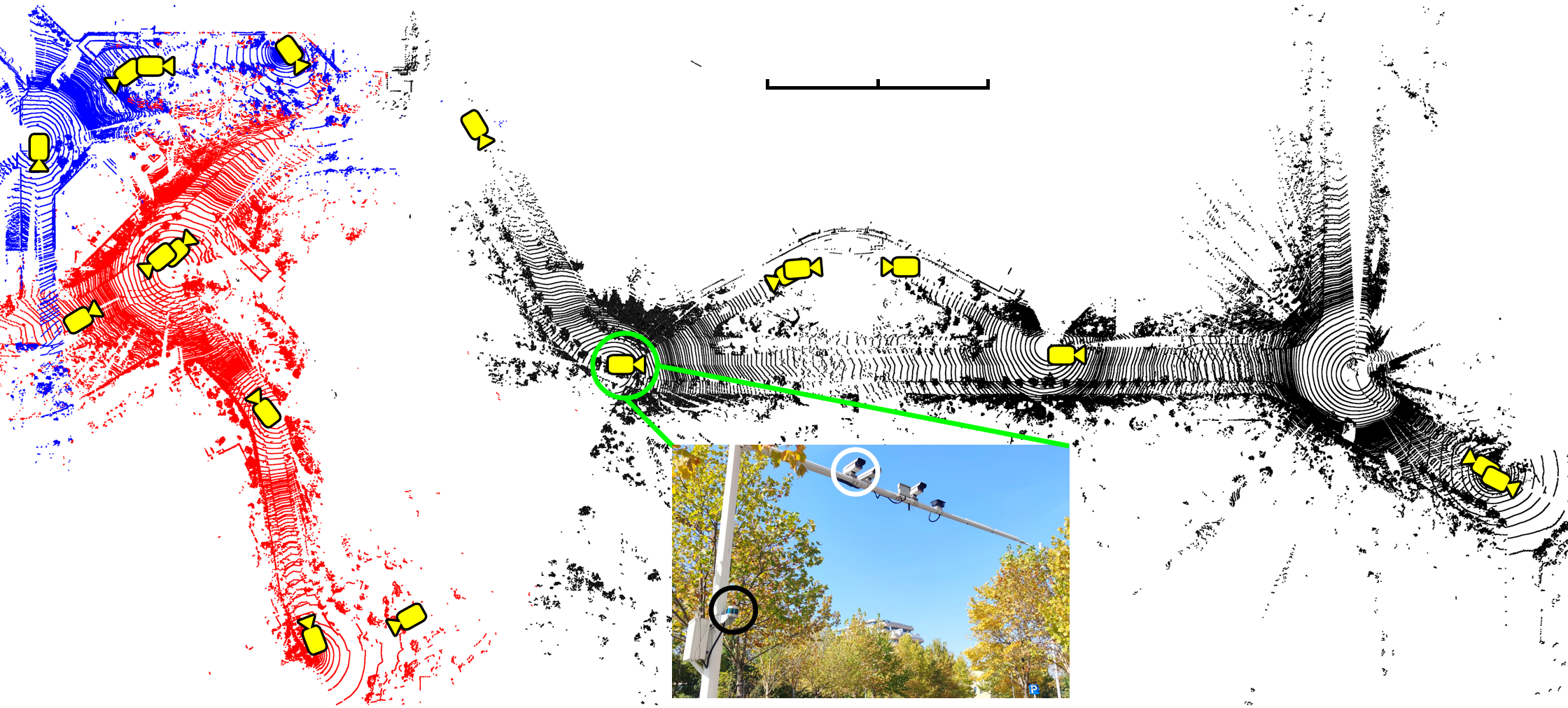}
\put(48.5, 41){0}
\put(55.0, 41){50}
\put(61.0, 41){100 m}
\put(5, 10){\color{blue}{\textbf{$\bigcdot$ A}}}
\put(5, 6){\color{red}{\textbf{$\bigcdot$ B}}}
\put(5, 2){\color{black}{\textbf{$\bigcdot$ C}}}
\put(48.5, 5.4){\colorbox{black}{\color{white}{\textbf{LiDAR}}}}
\put(50.0, 11.0){\colorbox{white}{\color{Black}{\textbf{Camera}}}}
\end{overpic}
\caption{A BEV overview of the data pattern in CORP. The colored are LiDAR point clouds and the schematic yellow shapes indicate the positions of cameras. The blue points cover a corner of the campus, with the red points monitoring two crossing paths and the black takes a wide area under surveillance. The three regions are denoted as A, B and C respectively, characterized by different detection ranges.}
\label{fig:CORP_overview}
\end{figure}
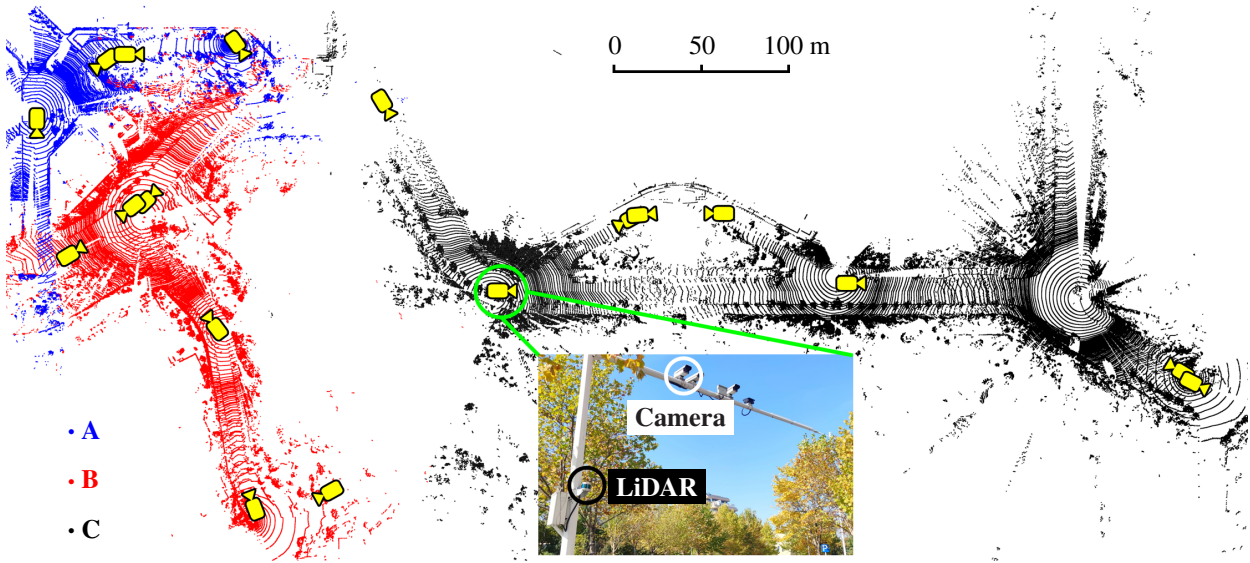

\begin{table}[tb]
\centering
\caption{A diverse sensor combinations deployed in the CORP dataset region-wise. For LiDARs, 32-b and 80-b stand for the number of laser beams. The ratio between the number of cameras and LiDARs is maintained to be 2:1 in each region.}
\label{tab:sensor_layout}
\setlength{\tabcolsep}{2.8mm}
\begin{tabular}{c|ccccc|cc}
\toprule
\multirow{2}{*}{Region} & \multicolumn{2}{c}{\# Cameras} & \multicolumn{2}{c}{\# of LiDARs} & \multirow{2}{*}{Total} & \multicolumn{2}{c}{\# of data frames} \\ \cline{2-5} \cline{7-8} 
                        & 1080P           & DCI-4K           & 32-b            & 80-b            &                                      & Image              & Point cloud             \\ \hline
A                       & 3               & 1                & 1               & 1               & 6                                    & 44.4K              & 22.2K                  \\ \hline
B                       & 3               & 3                & 1               & 2               & 9                                    & 79.6K              & 39.8K                  \\ \hline
C                       & 5               & 3                & 1               & 3               & 12                                   & 81.8K              & 40.9K                  \\ \hline
Total                    & 11              & 7                & 3               & 6               & 27                                  & 205.8K             & 102.9K                   \\
\bottomrule
\end{tabular}
\end{table}

The highlighted observation points towards a research frontier in which residential areas akin to university campuses provide distinctive research vantages and real-world deployment scenarios, thereby broadening the scope of roadside perception technology development beyond the traditional confines of urban streets. Motivated by this, we have innovatively constructed a roadside perception dataset named CORP (\textbf{C}ampus-\textbf{O}riented \textbf{R}oadside \textbf{P}erception), which is the first large-scale, multi-modal roadside dataset specifically designed for campus and large public spaces. Within the CORP dataset, in response to the unconstrained nature of the campus environment, multiple cameras and lidars with diverse specifications have been employed and installed at varying heights and locations on roadside poles across the campus, capturing rich scene information from various perspectives. \cref{tab:sensor_layout} shows that according to the number of sensors and the complexity of scenes, we have partitioned the CORP dataset into three regions, namely Region A, Region B, and Region C, as shown in \cref{fig:CORP_overview}. 

We anticipate that the CORP dataset will serve as a valuable resource for the research community in advancing various sensing methods within public residential spaces. Our key contributions can be summarized as follows:

\begin{itemize}

\item The first roadside multi-modal perception dataset collected in a large-scale campus environment, with a multitude of images, point clouds and precise annotations suitable for a variety of tasks.


\item The first dataset with pixel-level annotations for moving objects and seamless cross-device tracking of targets from roadside perspectives, offering resources for sensing tasks under residential settings.

\item We have experimented with baseline methods for various perception tasks, thereby identifying distinct challenges in roadside perception within public parks that differ from those encountered on urban arterial roads.
\end{itemize}

\section{Related Work}\label{sec:related_work}
\subsection{Roadside Perception Datasets}
\subsubsection{Urban Datasets.} There has been a growing release of roadside datasets in recent years. The IPS300+\cite{isp300_2022}, DAIR-V2X\cite{yu2022dairv2x} and Rope3D\cite{rope3d_2022_CVPR} datasets are among the first urban traffic data inventories, with accurate and consistent annotations for 2D and 3D targets including bounding boxes verified by geometric constraints, as well as occlusion levels and truncation states. A9 and A9-i datasets\cite{a9_2022,a9i_2023} focus on urban highway and intersection scenarios encompassing roadside cameras and LiDARs.

\subsubsection{Campus Datasets.} Datasets exist for perception tasks in campus settings. The ShanghaiTech\cite{shanghaitech} dataset is designed for anomaly detection with images, which contains videos taken from multiple scenes in a university in Shanghai, and hundreds of footages are labeled with various types of abnormal events. The NWPU Campus\cite{nwpu} is built for the same task on a larger scale. The Usyd Campus\cite{zhou2020USyd} dataset provides a multi-modal collection over and around the University of Sydney campus for a continuous period of 1.5 years, focusing on vehicle-side perception. The Campus3D\cite{Li2020Campus3D} dataset was generated through photogrammetric processing of UAV images taken over the National University of Singapore (NUS), which provides a dense point cloud of a campus tailored for point segmentation tasks. A comparison between CORP and the aforementioned datasets can be found in \cref{tab:dataset_comparison}.


\begin{table}[tbp]
\centering
\caption{A comparison between CORP and related datasets, encompassing the type (Tp) of the dataset with U standing for urban and C for campus, the number of point clouds (PCD), RGB and infrared (IR) images, the number of classes (Cls), the count of labeled 2D and 3D bounding boxes (Bx), the support for 2D segmentation (Seg) and seamless cross-device tracking (X-Trk). CORP displays dominance in data volume and extends roadside perception with pixel segmentation and cross-device tracking abilities.}
\label{tab:dataset_comparison}
\begin{threeparttable}

\begin{tabular}{c|c|c|c|c|c|c|c|c|c}

\toprule
Dataset  & Tp & Yr & \# PCD & \# RGB & \# IR & \# Cls & \# 2D/3D Bx & 2D Seg. & X-Trk\\
\hline
A9\cite{a9_2022}               & U & 2023 & 0.5k & 0.6k & \ding{56} & 9 & \ding{56}/14.5k & \ding{56} & \ding{56} \\
A9-i\cite{a9i_2023}            & U & 2023 & 4.8k & 4.8k & \ding{56} & 10 & \ding{56}/57.4k & \ding{56} & \ding{56} \\
IPS300+\cite{isp300_2022}      & U & 2022 & 14.2k  & 14.2k & \ding{56}& 7 & 4.54M/4.54M &\ding{56} & \ding{56} \\
Rope3D\cite{rope3d_2022_CVPR}  & U & 2023 & \ding{56} & 50k & \ding{56} & 12 & 670k/1.5M & \ding{56} & \ding{56} \\
DAIR-V2X\cite{yu2022dairv2x}   & U & 2023 & 71k & 71k & \ding{56} & 10 & 1.2M/1.2M & \ding{56} & \ding{56} \\
V2X-Seq\cite{v2xseq_2023}      & U & 2023 & 15k & 15k & \ding{56} & 10 & 0.33M/0.33M & \ding{56} & \ding{56} \\
INT2\cite{int2}                & U & 2023 & \ding{56} & \ding{56} & \ding{56} & 3 & \ding{56}/\ding{56} & \ding{56} & \ding{56} \\  
\hline
Shanghaitech\cite{shanghaitech} & C & 2016 & \ding{56} & 1198 & \ding{56} & 1 & 330k/\ding{56} & \ding{56} & \ding{56} \\  
USyd\cite{zhou2020USyd}         & C & 2020 & 300 & 300 & \ding{56} & 12 & \ding{56}/\ding{56} & \ding{52} & \ding{56} \\  
Campus3D\cite{Li2020Campus3D}   & C & 2020 & 937.1M & \ding{56} & \ding{56} & 24 & \ding{56}/\ding{56} & \ding{52} & \ding{56} \\  
NWPU\cite{nwpu}                 & C & 2023 & \ding{56} & 1466k & \ding{56} & 28 & \ding{56}/\ding{56} & \ding{56} & \ding{56} \\  
\hline
\textbf{CORP(Ours})             & C & /  & 102k & 205k  & 20k   & 5  & 296k/215k\tnote{*} & \ding{52} & \ding{52} \\

\bottomrule
\end{tabular}

\begin{tablenotes}
\footnotesize
\item[*] 20\% of the data frames are annotated.
\end{tablenotes}
\end{threeparttable}

\end{table}


\subsection{Object Detection and Tracking}
\subsubsection{2D object detection.} In the realm of 2D object detection based on images, deep learning methods have achieved significant performance improvements. Many recent developments in this field are focusing on enhancing the efficiency and speed of object detection algorithms for real-time applications. A substantial portion of recent developments in this field have focused on enhancing the efficiency and speed of real-time target detection algorithms. For instance, the YOLO series of models\cite{yolo_2016,yolo9000_2017,yolov3_2018,yolov4_2020,yolov5_2020,yolor_2021,yolov7_2022,yolov8_2023} have been widely adopted in industry due to their balance between efficiency and accuracy.Despite these advancements, issues such as occlusion and suboptimal lighting conditions continue to pose challenges in this domain. In this work, to assess the performance and robustness of 2D detectors in a campus environment, we employ YOLO-v5\cite{yolov5_2020} and v8\cite{yolov8_2023} for image-based detection.
\\
\subsubsection{2D Object Segmentation.} Segmentation of moving objects identifies image pixels belonging to foreground targets in motion or moving class instances from the background. Several motion segmentation benchmarks\cite{modnet_2018,instance_mot_seg_2021} have been on the shelf, with more recent segmentation methods\cite{learn_to_seg_2021, discover_move_2022, seg_move_2022} focusing on moving targets beyond the autonomous driving scenarios. These datasets and methods are built upon vehicle sensors and thus predominantly represent the environment in the car's view, while our CORP dataset supports motion segmentation tasks with real-world images from roadside perspectives.
\\
\subsubsection{3D detection.} 3D object detection involves a variety of methods including those based on point clouds, those based on images, and those combining both approaches. PointPillars\cite{pointpillars} encodes point clouds into pillar-shaped voxel features under Bird's Eye View (BEV), subsequent methods such as \cite{voxelnet_2018, 3dssd, voxelrcnn_2021, yin2021center} often perform detection tasks with point-wise, voxel-based or ROI-level feature representations. While point cloud detection delivers superior accuracy, camera-based methods \cite{neighborvote_21, zhang2021monoflex, rukhovich2022imvoxelnet, monoatt_cvpr23, unimode_cvpr24} are emerging for more efficient solutions. Recent works such as \cite{yang2023bevheight, monouni_nips23} are specially designed for roadside perception scenarios. Fusion-based methods enhance the robustness of detection with the fusion of modalities\cite{vora2020pointpainting, liu2023bevfusion, 3dcvf}. For example, BEVfusion~\cite{liu2023bevfusion} transfers the 2D camera image features and 3D LiDAR features to BEV space, allowing for a consistent alignment of features from varied viewpoints.
\\
\subsubsection{Object tracking.} 
Tracking is to achieve correspondence between new targets and historical data to identify their trajectories. For 2D tracking in images, SORT\cite{sort_2016} and DeepSORT\cite{deepsort_17} established a pipeline to incorporate Kalman filter and matching algorithms. OC-SORT\cite{ocsort_cvpr2023} extends SORT with directions of objects during similarity calculation. For 3D tracking, most methods also follow the tracking-by-detection paradigm, where the detector predicts target positions and categories, after which the tracking module generates continuous trajectories via association between consecutive frames. AB3D~\cite{ab3d} is a typical 3D tracking baseline method, which achieves the data association of detection results via the Hungarian matching algorithm with a minimal computation cost. More methods such as \cite{yin2021center, alphatrack} perform the data association with the motion or appearance features to enhance the tracking performance.

\section{CORP Dataset}\label{sec:corp}
CORP is the first campus-based, multi-modal roadside perception dataset that integrates natural and humanistic scenes, supporting tasks such as object detection, pixel-level segmentation of moving targets, and seamless remote tracking of objects across sensors. This dataset encompasses over 200,000 images and more than 100,000 point cloud data frames, all of which have been gathered through a collaborative effort by 9 LiDAR and 18 camera sensors strategically positioned in diverse and characteristic settings throughout the campus. This wealth of sensory data presents to the current roadside perception technologies a potentially valuable examination platform, which will be released soon with a development toolkit for data visualization, model training, and evaluation on our project webpage: \url{corp-dataset.github.io}.

\subsection{Sensor Configuration}
\subsubsection{Sensor Setup.} In the CORP dataset, the data frames were meticulously accumulated within a university campus, incorporating high-resolution imagery from 18 traffic surveillance cameras and spatial information captured by an ensemble of 9 automotive-grade LiDAR sensors. Each of these sensors was strategically installed atop intelligent poles at varying heights ranging from 2.5 to 6.5 meters above the ground, thereby ensuring comprehensive and diversified coverage for robust roadside perception tasks and presenting new challenges. An overview of the point cloud pattern and the position of cameras is shown in \cref{fig:CORP_overview} with a real-world picture of the pole. The range of the whole area extends about 700 meters in the west-east direction and has a north-south length of 350 meters. We divide the area into three regions, labeled A, B, and C, based on the detection range, and color-coded with blue, red, and black, respectively. We have configured different sensor layout schemes for different regions, and the specific number and distribution details of the sensors in the layout are shown in Table~\ref{tab:sensor_layout}.


\subsubsection{Coordinate Systems.} 
\begin{figure}[t]
\centering
\includegraphics[width=1.0\columnwidth]{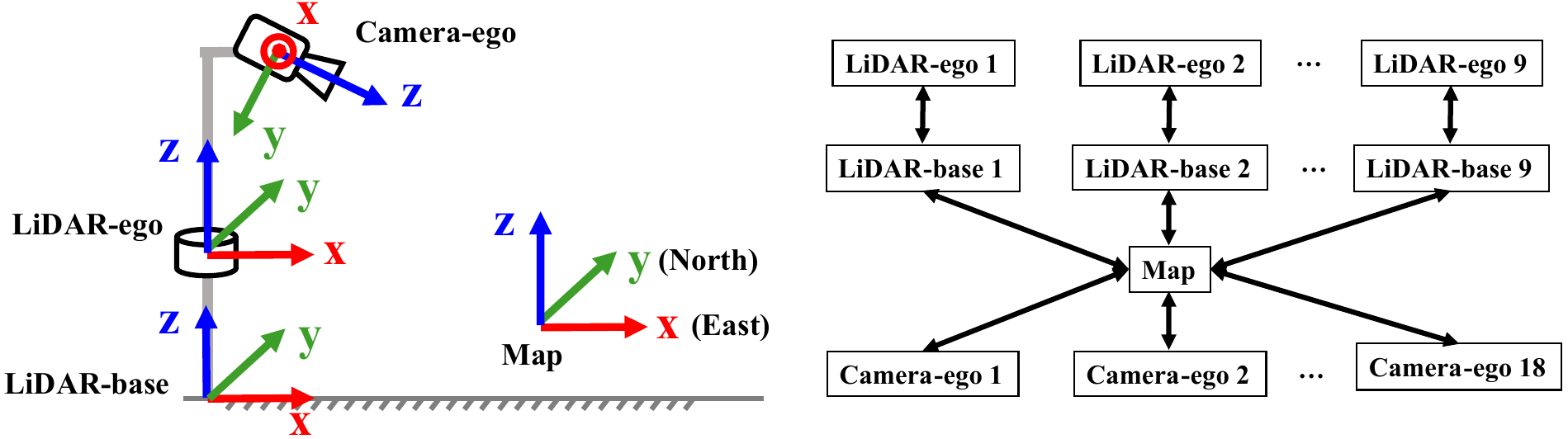}
\caption{An overview of 4 types of coordinate systems involved in CORP, \ie Camera-ego, LiDAR-ego, LiDAR-base and Map systems. On the right side is an illustration of their calibration relationships, where double-headed arrows denote the presence of the calibration parameters.}
\label{fig:coord_system}
\end{figure}

We have constructed four types of coordinate systems for the CORP dataset, of which two are used for LiDAR sensors, one for cameras, and the last one for the global map, as illustrated in \cref{fig:coord_system}. For LiDARs, the LiDAR-ego reference frame originates from the sensor itself with the $x$-axis pointing forward and $z$-axis upward. For the convenience of researchers and to safeguard data privacy, we have constructed an additional virtual LiDAR coordinate system, termed as LiDAR-base, transforms the origin of LiDAR-ego to the ground contact point of its host pole to comply its $x$-$o$-$y$ plane with the local ground surface. For cameras, the origin of the camera-ego coordinate system overlaps the optical center of the sensor with the $z$-axis pointing forward along the camera's view and $x$-axis pointing leftward accordingly. The LiDAR and camera coordinate systems are employed to represent objects in the sensor's field of view, while the map coordinates unify the objects into a global picture. The map coordinate system has an origin with known GPS,  its $x$-axis coincides with the east and the $y$-axis with the north.

\subsubsection{LiDAR Calibration.} With a default LiDAR-ego coordinate system, a LiDAR-base is calibrated automatically. To be specific, the LiDAR point clouds fell on the ground plane approximately form small-range concentric circles, therefore one is able to identify the axes of the projected LiDAR-base system and solve for the extrinsic transformation. Further, using a point cloud map of the area taken by an unmanned aerial vehicle, we are able to calibrate the LiDAR-base coordinates to the global map system.

\subsubsection{Camera Calibration.} Intrinsic parameters of a camera define the projection between the incoming rays and image pixels. For CORP, we use chessboards as explicit targets and employ MATLAB Camera Calibrator \cite{CameraCalibrator} following the method outlined by Zhang \etal\cite{zhang2000flexible} to determine the intrinsic calibration. This procedure yields a set of quantities including $[f_x, f_y, c_x, c_y, k_1, k_2, p_1, p_2, k_3]$, where $f_x$ and $f_y$ denote the focal lengths, while $c_x$ and $c_y$ represent the principal points. Additionally, the parameters $k_1$, $k_2$, and $k_3$ pertain to radial distortion, and $p_1$ and $p_2$ are related to tangential distortion.


Camera extrinsic parameters are calibrated by optimizing the camera's position and orientation in favor of the similarity between the image and the colored point cloud map in the camera's FOV. Specifically, starting from the point clouds and a coarse camera pose, we generate a high-resolution image of the map in the camera's view by employing an interpolation algorithm. Subsequently, we apply a feature matching algorithm between the generated image and its RGB counterparts to remove outliers to refine the pose of the camera, and then the re-projection error between the 2D-3D feature pairs is minimized to achieve a better calibration. Camera-ego coordinates are aligned to the origin of the point cloud map same as LiDARs, with which one can easily find the coordinate transformation between any camera and LiDAR.

\subsection{Data Acquisition and Annotation}
\begin{figure}[tb]
\centering
    \begin{minipage}[b]{.325\columnwidth}
        \centering
        \includegraphics[width=1.0\columnwidth, height=0.56\columnwidth]{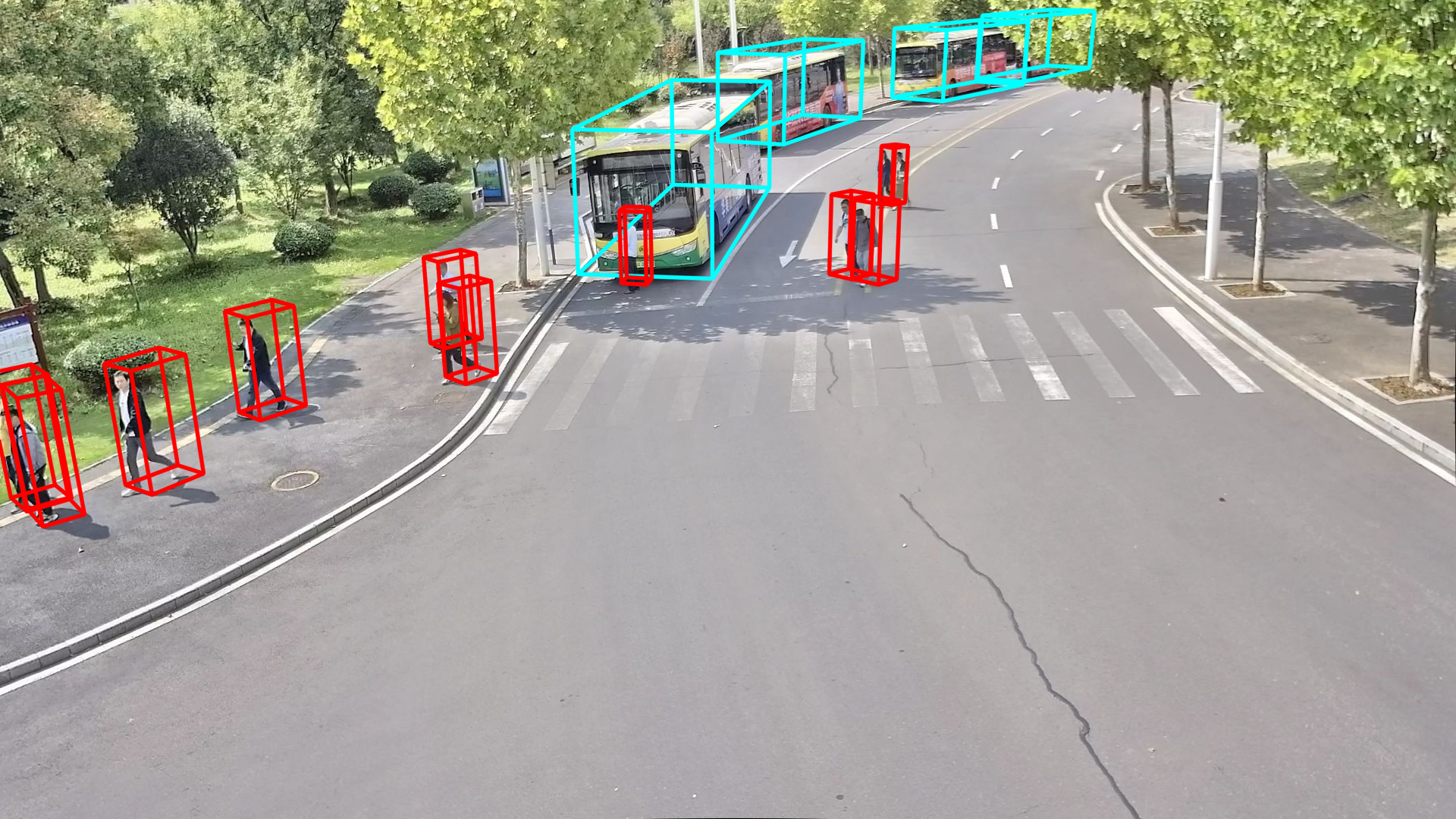} \\
        \centering
        \includegraphics[width=1.0\columnwidth, height=0.56\columnwidth]{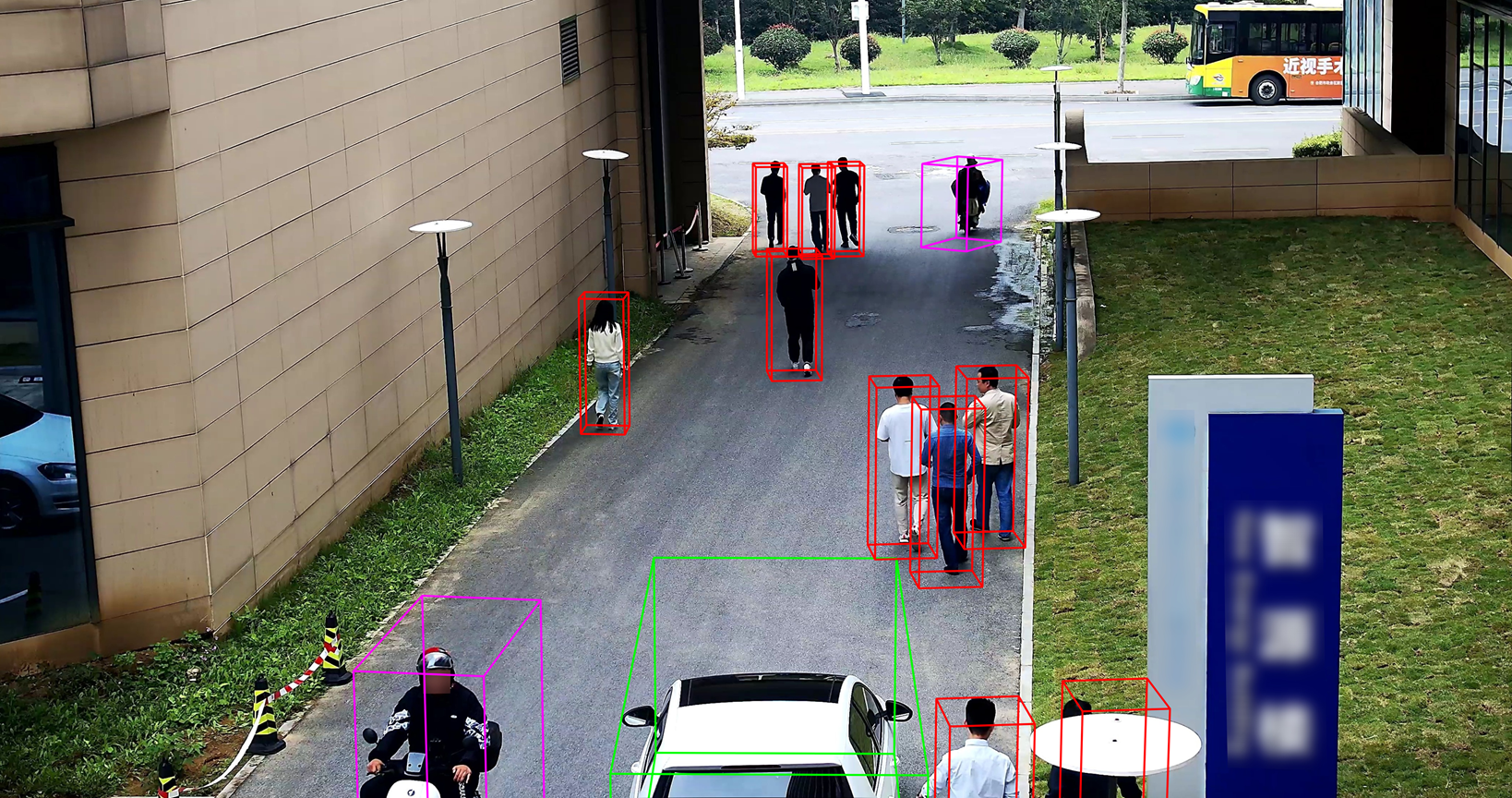} \\
        \centering
        \includegraphics[width=1.0\columnwidth, height=0.56\columnwidth]{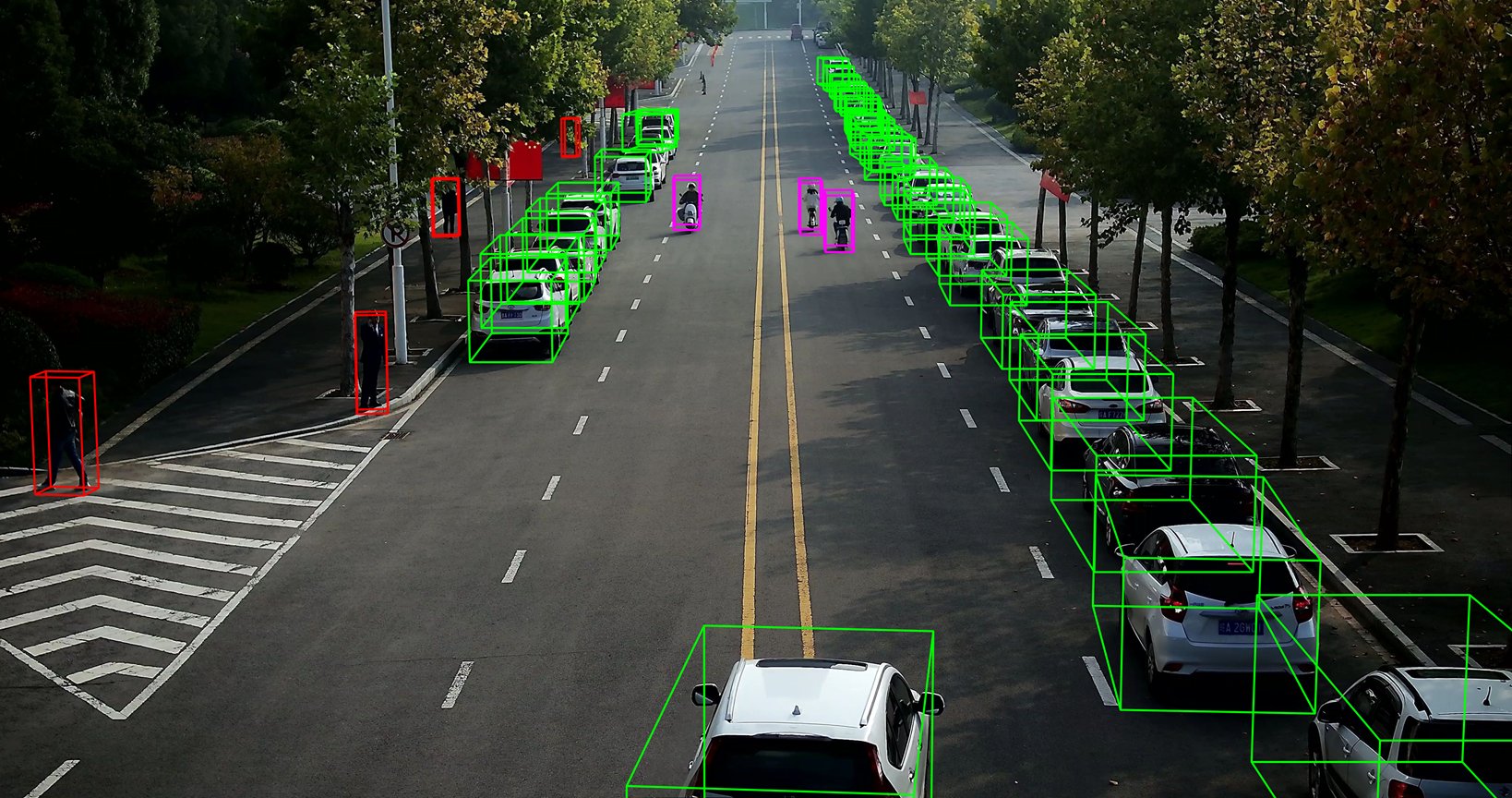}\\
        \subcaption{3D boxes}
    \end{minipage}\hspace{0pt}
    \begin{minipage}[b]{.325\columnwidth}
        \centering
        \includegraphics[width=1.0\columnwidth, height=0.56\columnwidth]{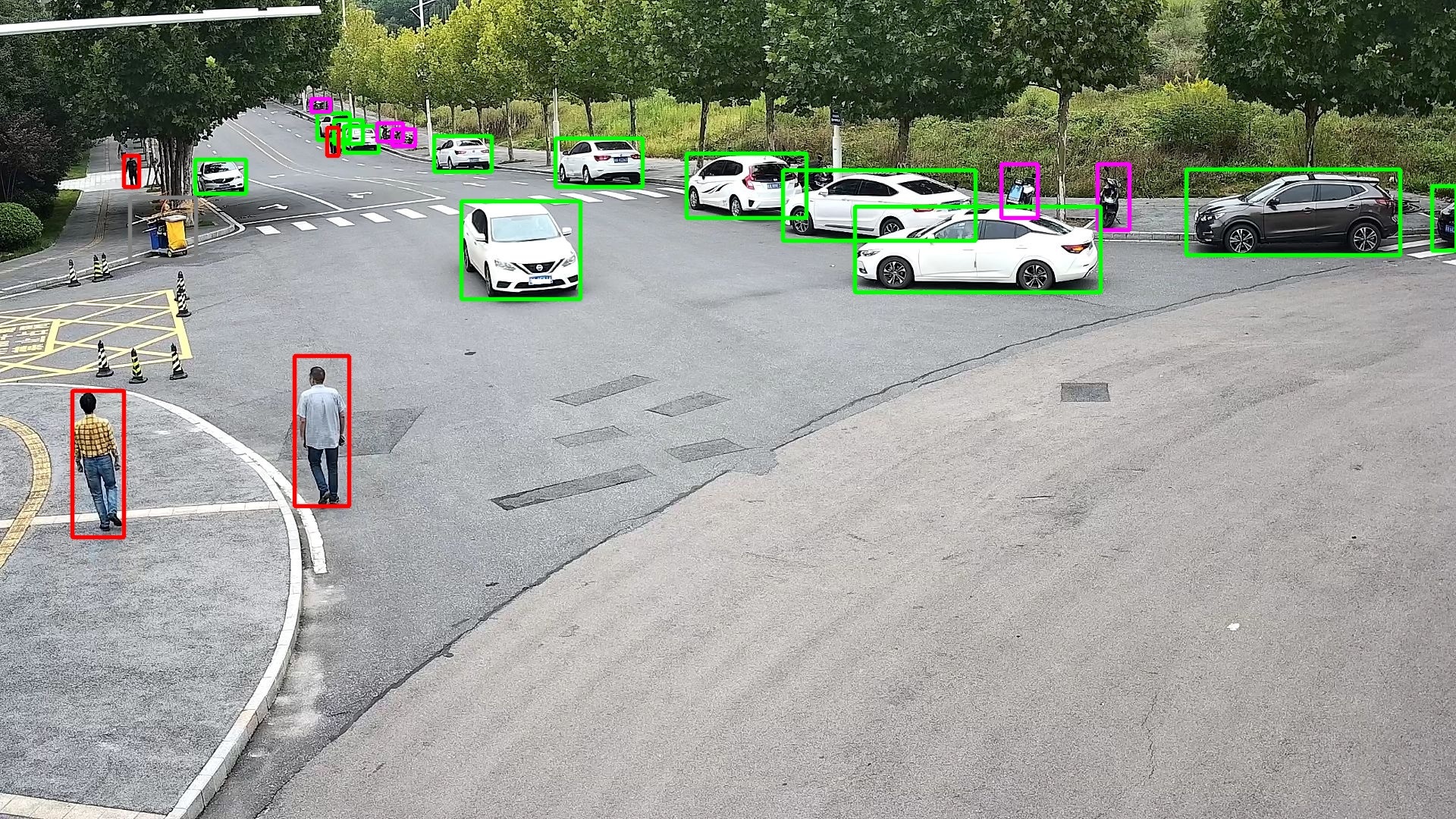} \\
        \centering
        \includegraphics[width=1.0\columnwidth, height=0.56\columnwidth]{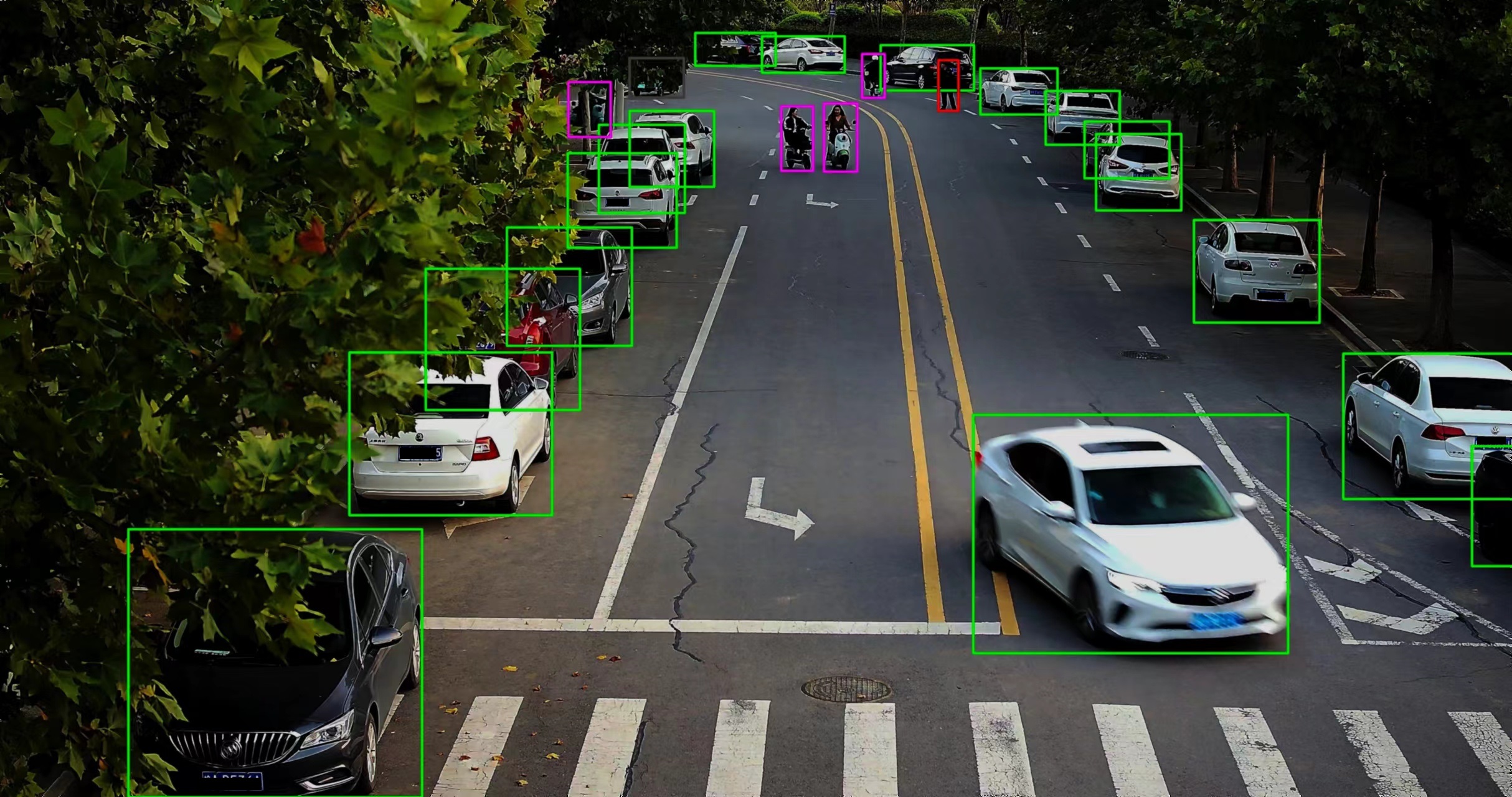} \\
        \centering
        \includegraphics[width=1.0\columnwidth, height=0.56\columnwidth]{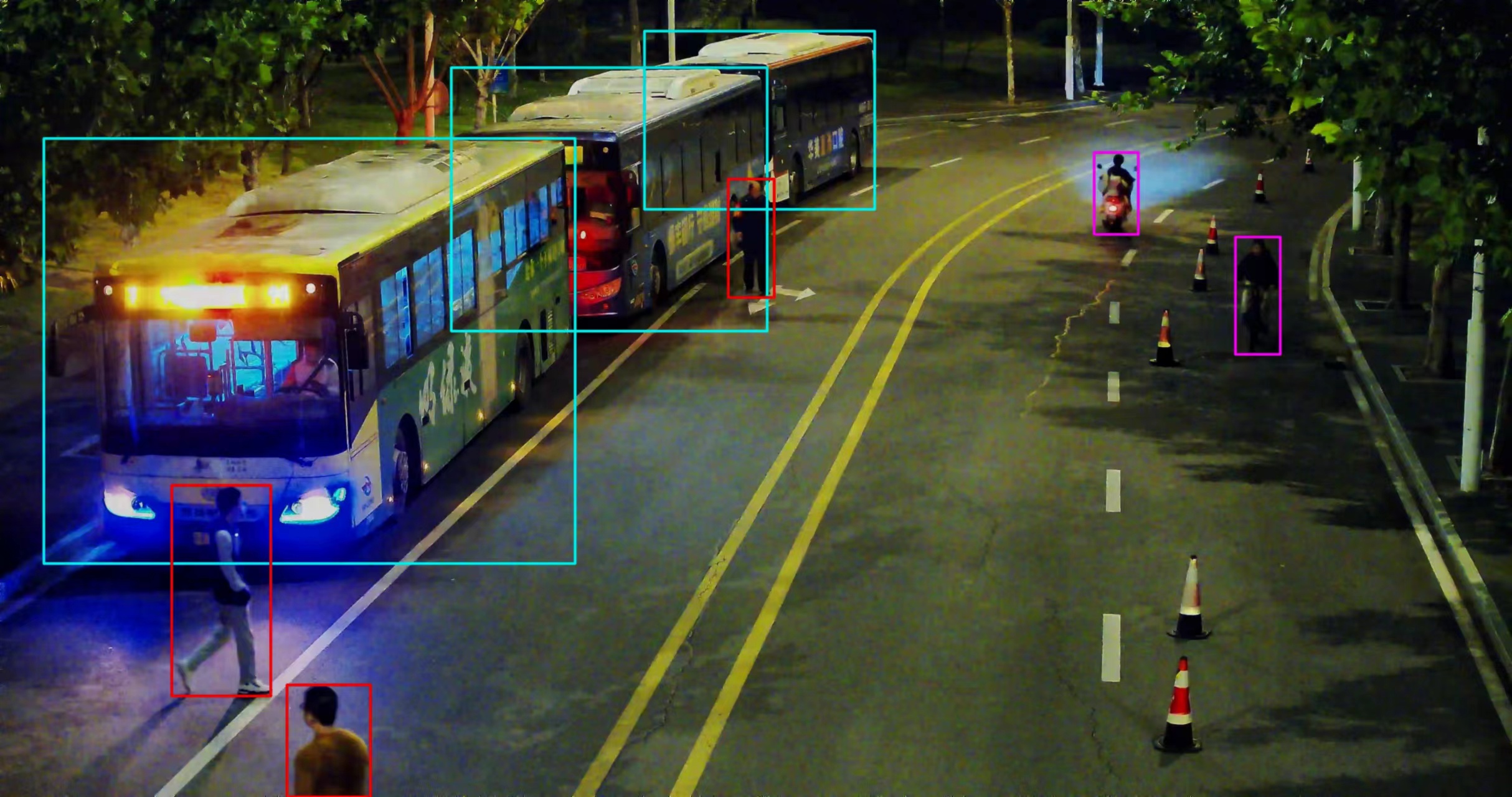}
        \subcaption{2D boxes}
    \end{minipage}\hspace{0pt}
    \begin{minipage}[b]{.325\columnwidth}
        \centering
        \includegraphics[width=1.0\columnwidth, height=0.56\columnwidth]{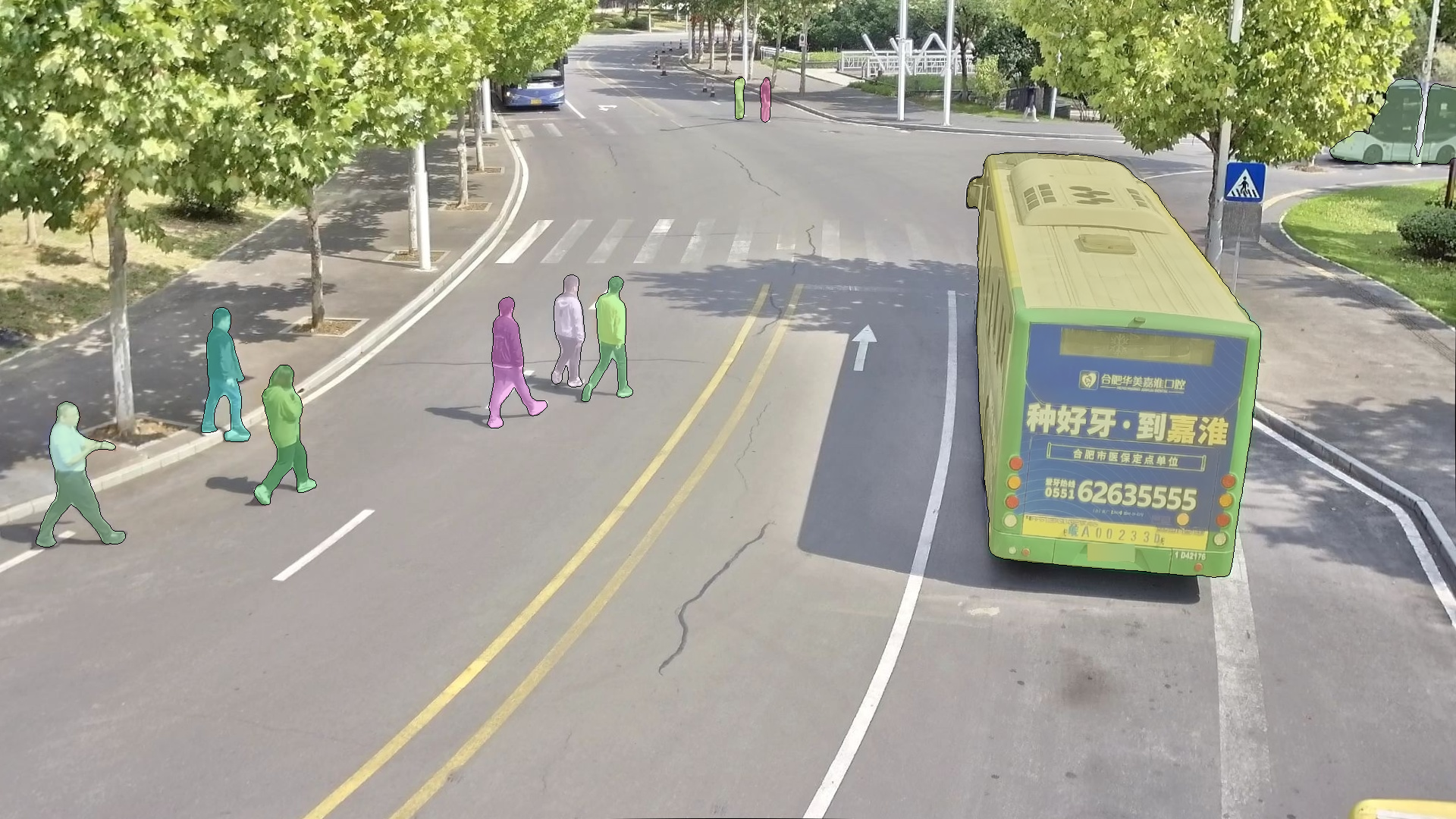} \\
        \centering
        \includegraphics[width=1.0\columnwidth, height=0.56\columnwidth]{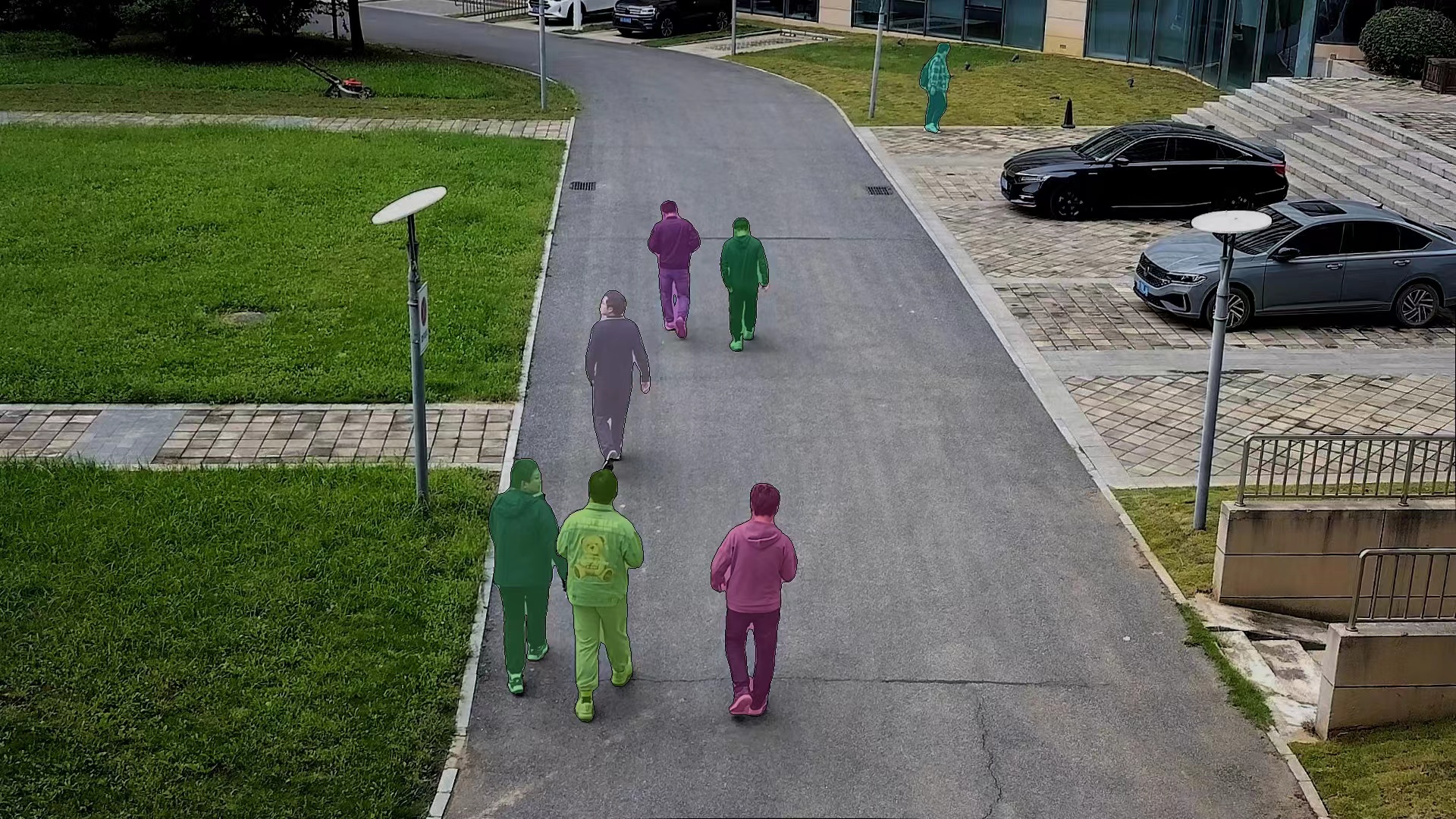} \\
        \centering
        \includegraphics[width=1.0\columnwidth, height=0.56\columnwidth]{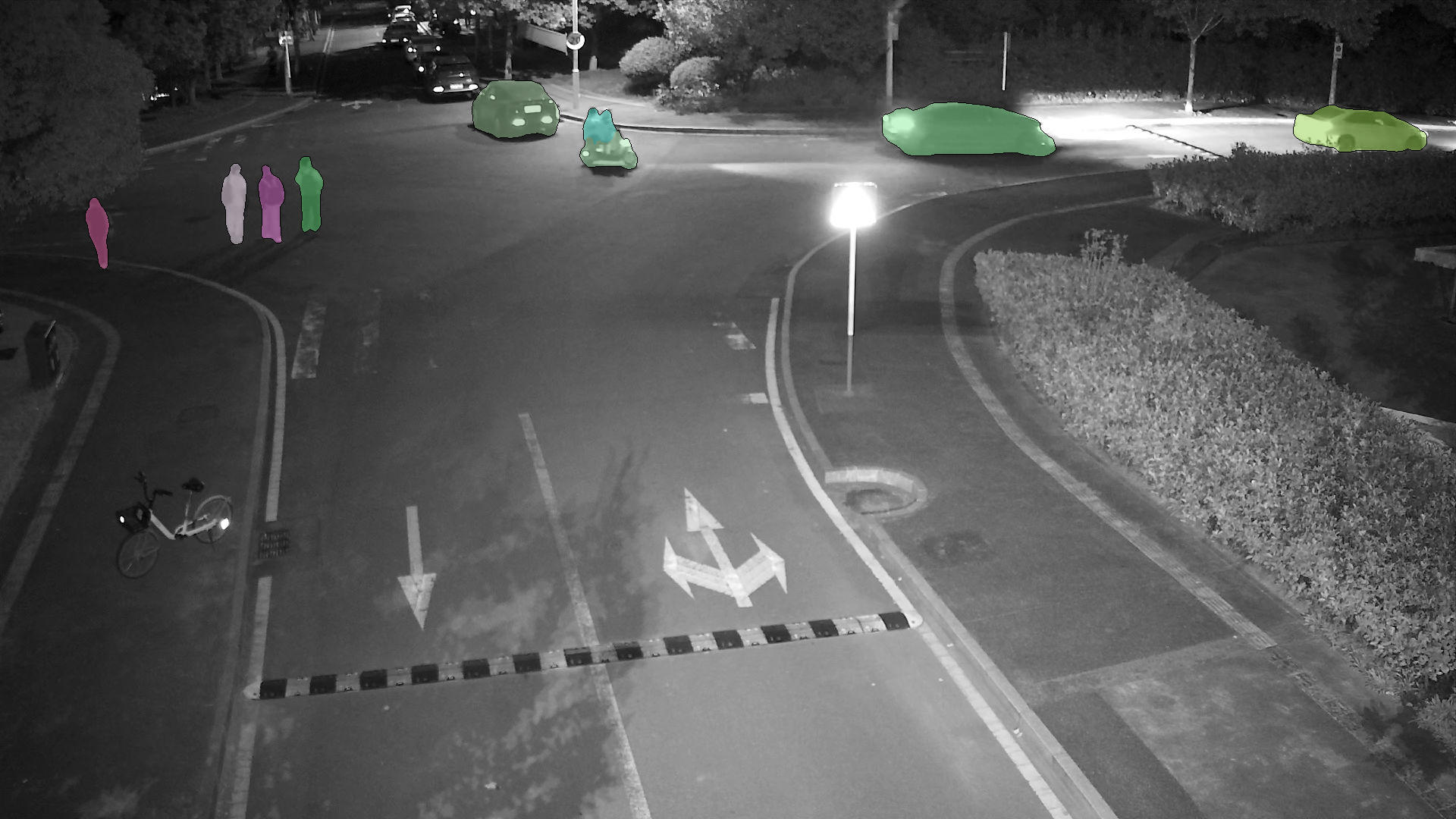}
        \subcaption{2D masks}
    \end{minipage}
\caption{CORP image examples with (a) 3D annotations, (b) 2D boxes, and (c) segmentation masks. All are images taken from different cameras, with illumination diminishing from the top, and note that each image has all of the mentioned annotations.}
\label{fig:examples}
\end{figure}
\noindent\textbf{Data Collection.} To build CORP, images are captured at a rate of 20 Hz while maintaining GPS time synchronization. Simultaneously, the LiDAR data is acquired at 10 Hz with the same source for recording timestamps. Each episode of the time sequence data in CORP lasts for 30 to 32 seconds with image and LiDAR frames from multiple sensors, the quantity of which varies across different regions. Subsequent to data collection, a precise alignment between the camera and the LiDAR timelines is performed in a post-processing manner with a tolerance of 20 milliseconds, to arrive at a final frame rate of 10 Hz for the data episodes. For scene diversity and considering the specificity of the campus environment, we select the data based on sparsity of the scene and get an average of 10 targets per LiDAR frame. Moreover, around 20\% of the data frames are recorded in the evening and 80\% in the daytime including morning, noon, and dusk with clear and cloudy weather conditions. The diversity within the CORP dataset would enable the robustness of perception algorithms to various campus-specific backgrounds and lighting conditions.\\
\\
\noindent\textbf{Data Labeling and Visualization.} Human annotators are employed for joint labeling of the 2D and 3D targets in the dataset using the format in DAIR-V2X, with additional information such as target ID and motion status (\textit{moving}v.s. \textit{static}) added. To generate pixel segmentation masks for the moving instances in mages, the foundation vision model SAM\cite{kirillov2023segment} is applied with the integration of human intervention and manual guidance. Visualization of data samples is shown in \cref{fig:examples}. The images and point clouds can also be viewed with their annotations from SUSTechPoints\cite{sustechpoints} with our development toolkit.

\begin{figure}[tb]
\centering
\includegraphics[width=1.0\columnwidth]{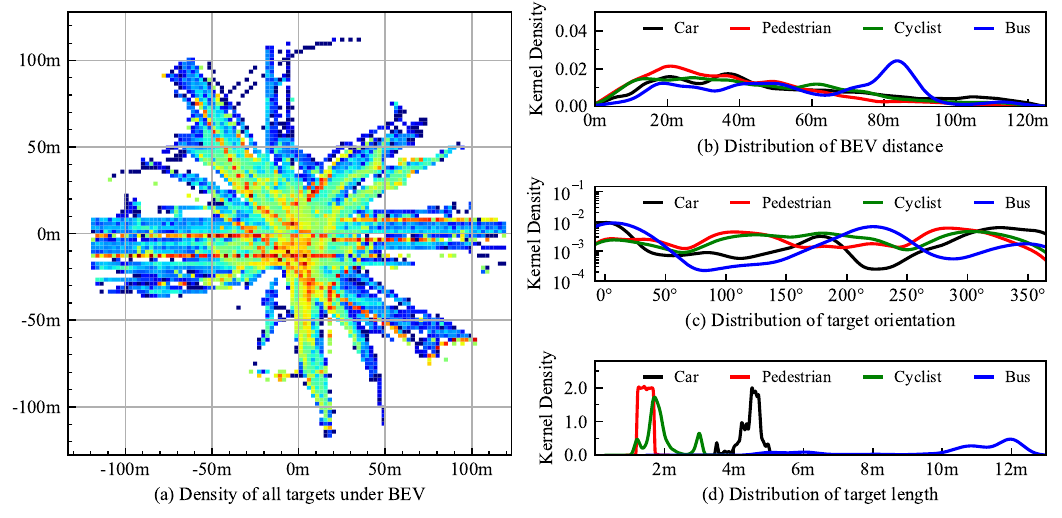}
\caption{Distribution of target location and orientation. (a) is a stacked overview of targets under BEV in their LiDAR-base systems, (b) and (c) are the distributions of their BEV distance and orientation relative to LiDARs, (d) represents the distribution of targets length in the dataset.}\label{fig:bev_distribution}
\end{figure}

\subsection{Supported Tasks and Metrics}
Equipped with ground-truth bounding boxes, jointly-annotated target IDs, and segmentation masks, our CORP dataset enables various roadside perception tasks such as 2D detection and 3D multi-modal perception, local and distant multi-objects tracking across devices, as well as pixel segmentation of moving targets. For the detection methods, we use Precision (P), Recall (R), and Average Precision (AP) as principle metrics for performance comparison. Similar to \cite{v2xseq_2023}, Multi-Object Tracking Accuracy (MOTA), and Multi-Object Tracking Precision (MOTP) are used for testing 3D tracking performance. For 2D moving instance segmentation, following \cite{wcj_u2onet_2021} we use P, R, IoU, and F-measurement for quantitative assessment.

\subsection{Features and statistics}
The CORP dataset encompasses 5 types of targets in a campus environment, including \textit{Bus}, \textit{Car}, \textit{Cyclist}, \textit{Pedestrian} and \textit{Other}, where the \textit{Other} category encompasses the tricycles used by janitors and carriers. The dataset is subjected to anonymization procedures in adherence to pertinent laws and regulations, thereby safeguarding privacy.

To get an overview of how targets are distributed in space, we aligned all identified 3D objects into their own LiDAR-base coordinate system, each of which has $x$-axis pointing eastward and $y$-axis towards the north, and visualized them collectively as shown in \cref{fig:bev_distribution}(a), where warmer color indicates a higher concentration of targets. Also, it is not surprising that the spatial distribution of targets exhibits a discernible trend toward specific directions as also demonstrated in their orientation profile in \cref{fig:bev_distribution}(c), since the mobility of targets is partially subjected to the road network in the environment. The allocation of targets indicates that our dataset encompasses plenty of road arrangements. \cref{fig:bev_distribution}(b) showed the distribution profile for the distance between targets and their center LiDARs under BEV. The portion of targets decreases as the distance increases since point clouds are sparser there, but most of the targets appear relatively evenly within 100 meters away from the center.


\begin{figure}[tb]
\centering
\includegraphics[width=1.0\columnwidth]{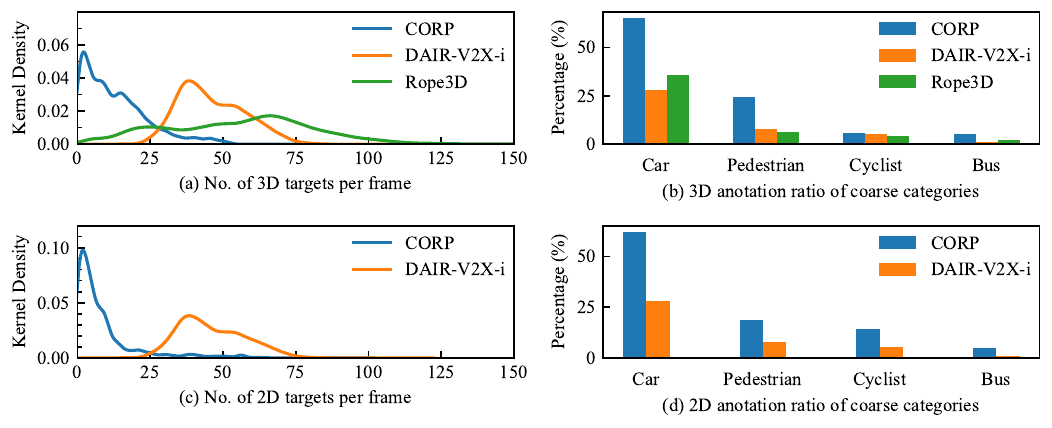}
\caption{A comparision of the target density in CORP and two of its urban-road counterparts. (a) and (c) are the distribution of per-frame 2D and 3D target count, respectively; (b) and (d) illustrates the percentage of 4 common types of road users,\ie Car, Pedestrian, Cyclist and Bus. Rope3D is not present in (c) and (d) since it has no 2D bounding boxes.}
\label{fig:data_compare}
\end{figure}

To glance at how targets distribute differently in CORP than in urban arterial streets, we analyzed the target densities of CORP, DAIR-V2X-i\cite{yu2022dairv2x} and Rope3D\cite{rope3d_2022_CVPR}. As shown in \cref{fig:data_compare}(a) and (c). CORP displays a sparsity of per-frame target numbers in both 2D and 3D cases, about half of the frames contain less than 15 targets while the other half has up to 50 targets per frame, which is not the case in its urban counterparts. However, the sparsity of targets does not imply simplicity of perception tasks, since accurately and efficiently localizing them from the background requires extra engineering for the model. In addition, 

Although the volume of the object class is less than that of DAIR-V2X and A9, we found no significant imbalance in the number of targets, indicating a relatively minor disparity over categories, thus diminishing the risks of label inequity. Note that, Rope3D has no 2D bounding box annotations and therefore is excluded in \cref{fig:data_compare}(c) and (d).


\section{Experiments}\label{sec:experiments}
With the CORP dataset, we perform experiments with methods widely recognized by the community on various tasks, including 2D detection and segmentation with images, 3D detection with cameras and LiDARs, 3D persistent tracking across devices, as well as non-learning methods for distance estimation. 
\subsection{Baseline Methods}
\subsubsection{2D Detection and Segmentation}
For 2D detection tasks, our baseline methods are YOLO-v5\cite{yolov5_2020} and the up-to-date YOLO-v8\cite{yolov8_2023} with input image resolution set to $1536\times864$. They were trained on CORP with 2 NVIDIA RTX-3090 GPUs for 100 epochs with SGD optimizer at a learning rate of 0.01. For motion segmention task, we applied a revised version of RiWNet\cite{wang2021riwnet}, termed as RiWNet+, for accurate and real-time segmentation of moving instances in a campus setting. Details of RiWNet+ will be released and open-sourced in a separate work.

\subsubsection{3D Detection and Tracking}
For LiDAR-based baseline, we trained PointPillars with a detection range customized to [-102.4, -102.4,-2.0, 102.4, 102.4, 4.0] and the size of the voxel grid as [0.16, 0.16] along $x$ and $y$ axes. All point clouds are transformed with annotations to the LiDAR-base coordinate system mentioned in \cref{sec:corp}. For camera-based detection, we compared the performance of BEVHeight and ImVoxelNet. In addition, we trained BEVFusion following \cite{liu2023bevfusion} but confined to perception sites with both camera and LiDAR installed. We adapt AB3D to the detection results as our tracking baseline.

\subsubsection{Per-object Distance Estimation}
In addition detection and tracking, we propose to rethink the distance estimation of targets w.r.t the camera, which is a very important application for self-driving and intelligent transportation systems\cite{vehicle_distance_wacv20, vechile_ipm_19, joint_det_and_dist_20}. Our motivation is two-fold: firstly, although 3D detection models are apparently capable of estimating the distance of targets, their generalizability is often ristricted\cite{eccv_2020_mono3d_general, li2022model_generalization}. Secondly, the large-scale deployment of learning-based detection methods is contingent upon the availability of substantial volumes of data, which is expensive and scarce.

Therefore, we propose a simple learning-free distance estimator leveraging the pin-hole camera model as our baseline. Specifically, with a known pose of the camera and its intrinsic parameters, the grounded point of targets in the image can be reconstructed as a 3D point in the camera-ego system. This method is termed as Pseudo-3D (P3D) in our experiments, more details of which are provided in the supplementary material. Our CORP dataset encompasses 18 cameras, each characterized by unique heights, pose angles and intrinsics, making it a suitable testbed for per-object distance estimation methods. 


\subsection{Results}
\subsubsection{Objection detection and Tracking.} 
\begin{figure}[tb]
\centering
    \begin{minipage}[b]{1.0\columnwidth}
        \centering
        \includegraphics[width=0.32\columnwidth, height=0.18\columnwidth]{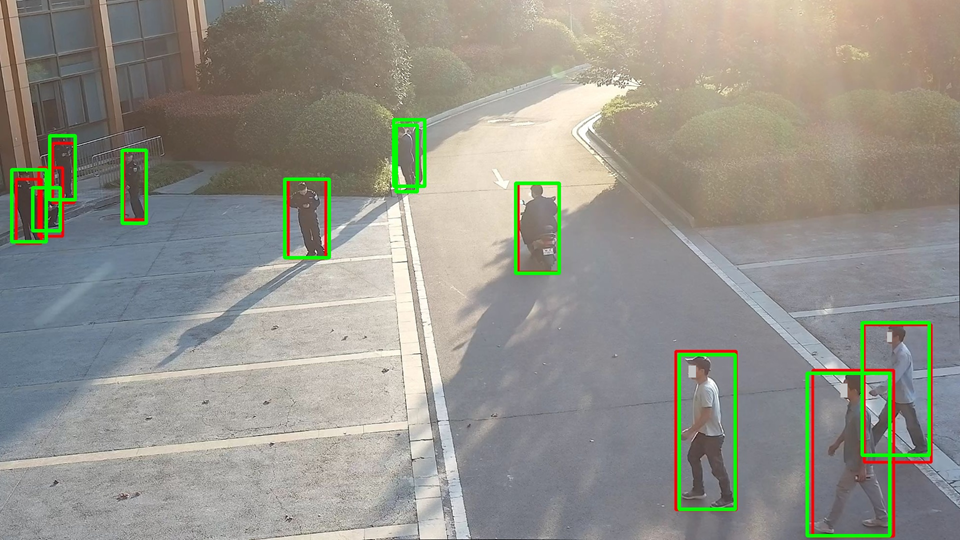}
        \includegraphics[width=0.32\columnwidth, height=0.18\columnwidth]{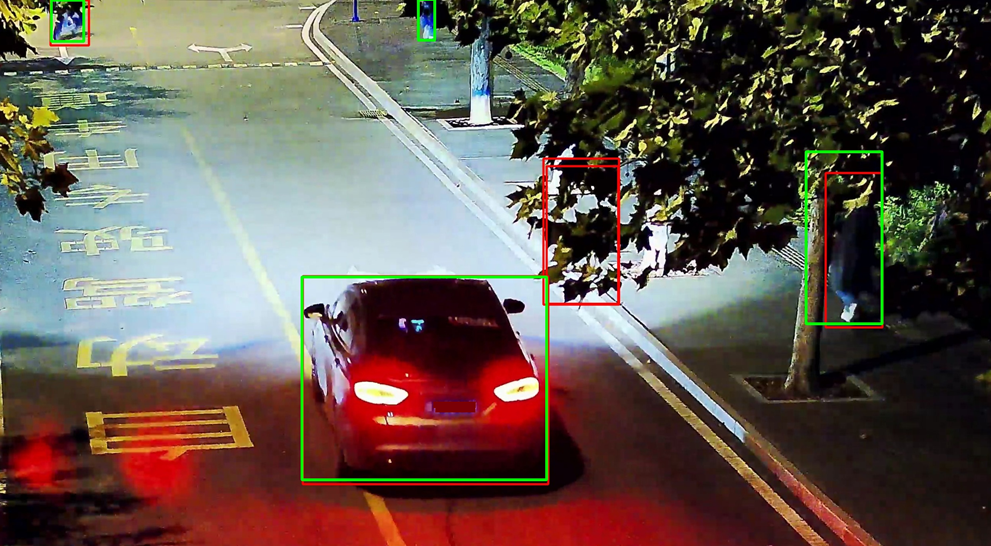}
        \includegraphics[width=0.32\columnwidth, height=0.18\columnwidth]{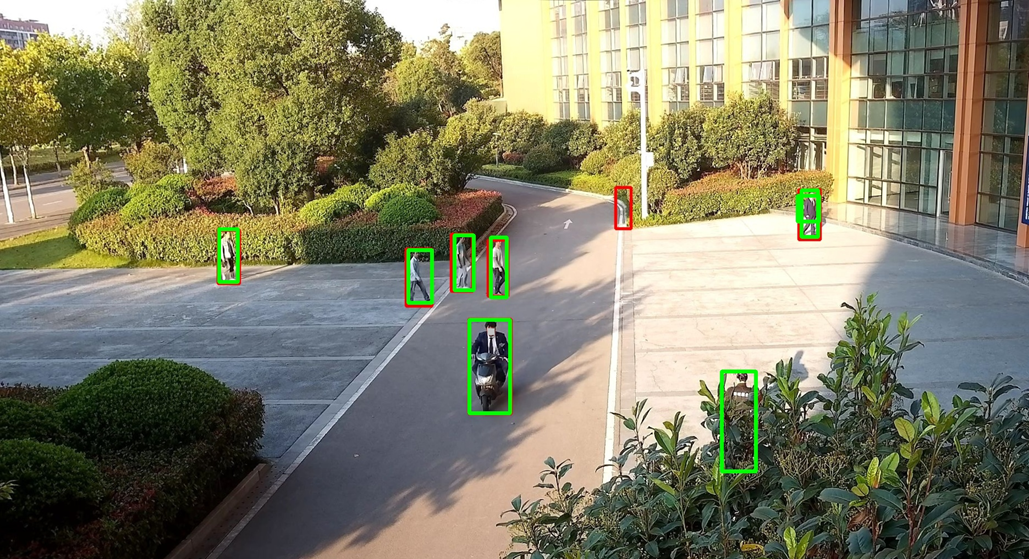}
        \subcaption{Qualitative detection results with YOLO-V8 on (from left to right): backlit scenes, occlusion by trees, contrast of lights}
    \end{minipage}\hspace{0pt}
    \\
    \begin{minipage}[b]{1.0\columnwidth}
        \centering
        \includegraphics[width=0.32\columnwidth, height=0.18\columnwidth]{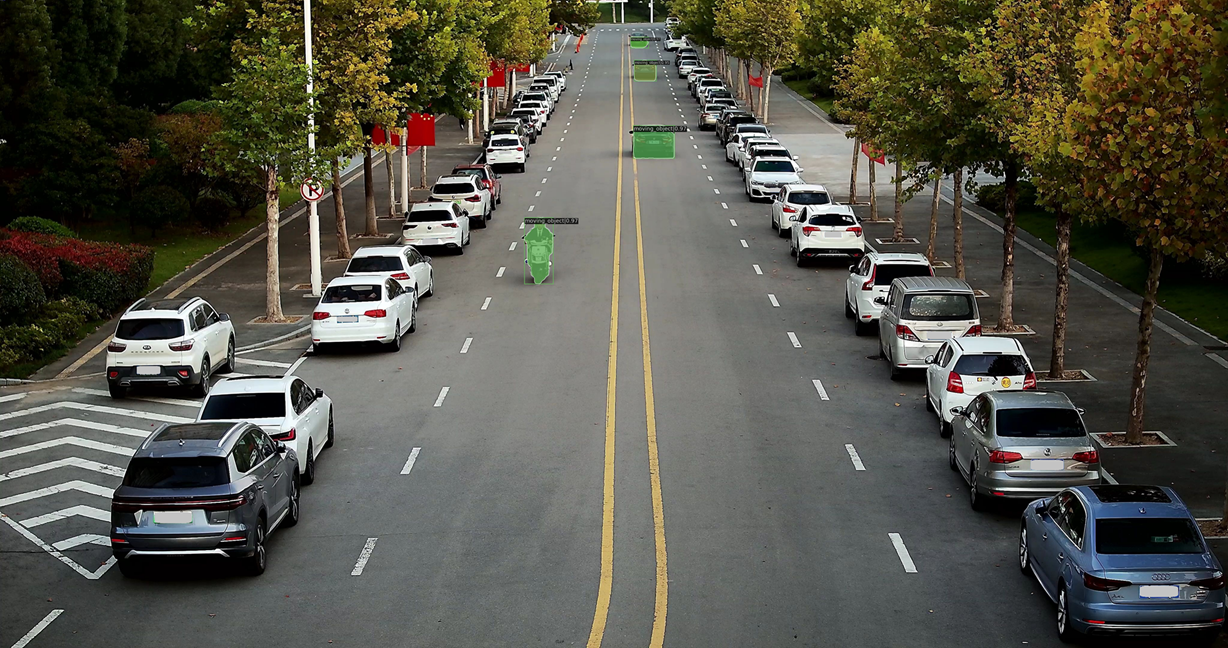}
        \centering
        \includegraphics[width=0.32\columnwidth, height=0.18\columnwidth]{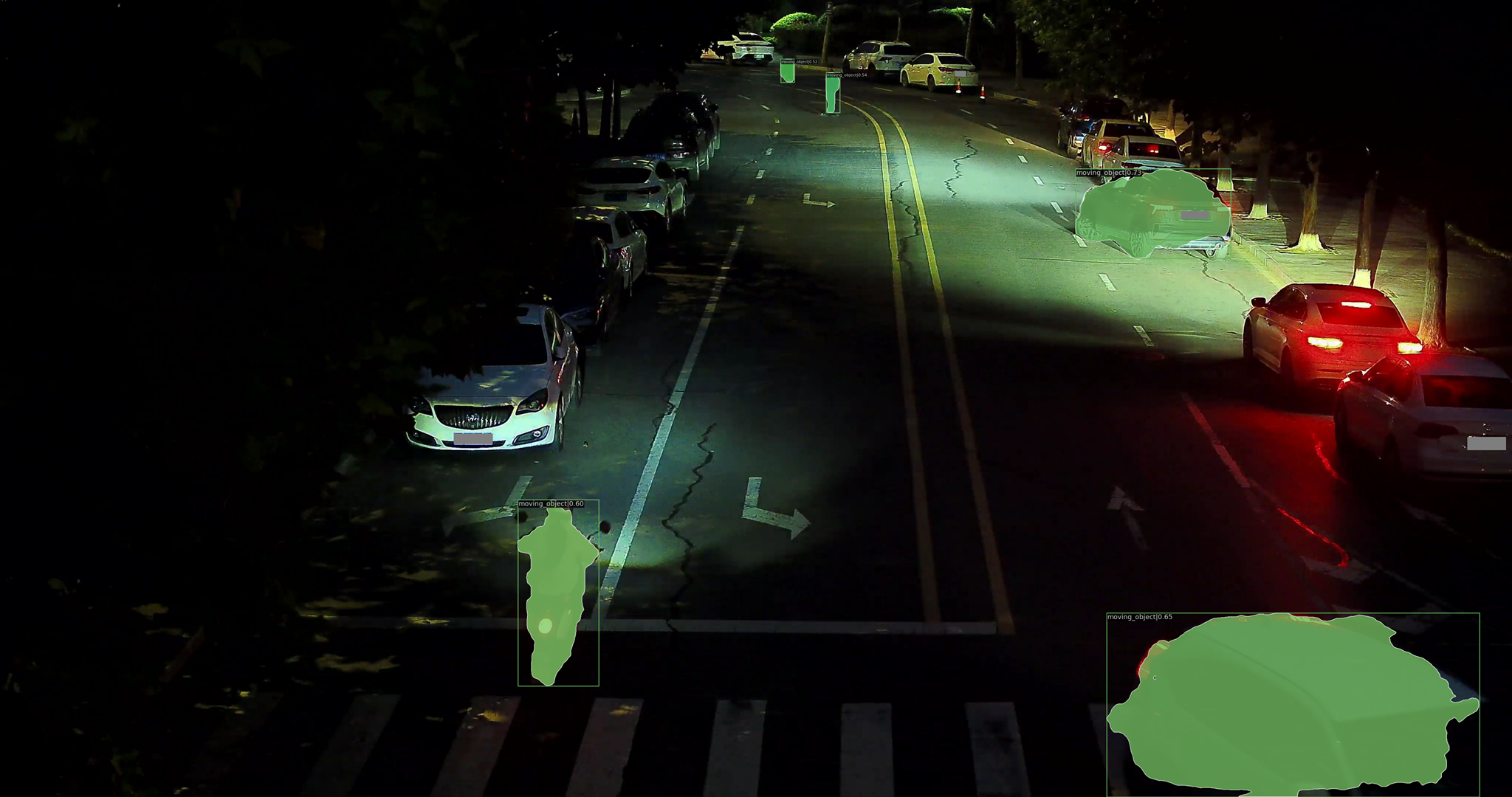}
        \centering
        \includegraphics[width=0.32\columnwidth, height=0.18\columnwidth]{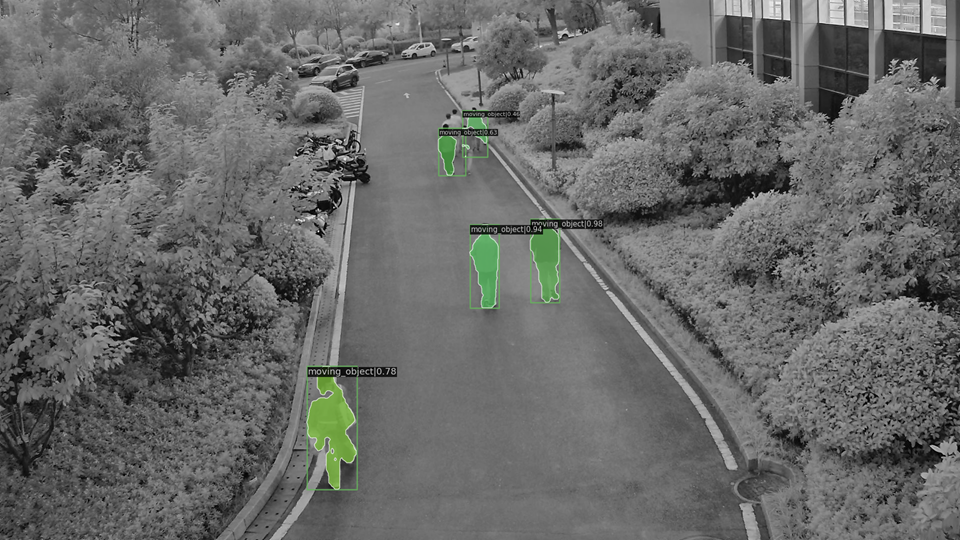}
        \subcaption{Qualitative moving instance segmentation results with RiWNet+ on (from left to right): mostly-static scenes, extreme low-light scenes, infrared images.}
    \end{minipage}\hspace{0pt}
\caption{Some challenging scenarios in CORP for object detection and segmentation tasks.}
\label{fig:challenging_examples}
\end{figure}

For 2D detection with CORP, we compared the performance of broadly acknowledged YOLO-V5 and its V8 counterparts. As shown in Table \ref{tab:detection2d}, the two methods demonstrate an obvious gap in overall precision, recall and AP scores. Despite the higher precision for YOLO-V5, V8 wins in terms of $AP_{50-95}$, implying a better generalization ability. However, challenges persist in accurately detecting distant targets, particularly when they are occluded by foreground elements or under suboptimal light conditions as shown in \cref{fig:challenging_examples}(a).

\begin{table}[tb]
    \centering
    \begin{minipage}[b]{0.48\columnwidth}
        \centering
        \caption{Evaluation of 2D detection methods.}
        \label{tab:detection2d}
        \footnotesize
        \begin{tabular}{c|c|c|ccc}
        \Xhline{0.6pt}
        \multirow{2}{*}{Method} & \multirow{2}{*}{P} & \multirow{2}{*}{R} & \multicolumn{3}{c}{AP$_{50-95}$} \\
        &       &       &  Veh.   & Cyc.    & Ped.     \\ \Xhline{0.6pt}
        YOLO-v5\cite{yolov5_2020} & 90.5 & 56.7 & 61.4  & 43.2   & 33.8    \\
        YOLO-v8\cite{yolov8_2023} & 87.4 & 61.2 & 71.1  & 50.5   & 39.1    \\ \Xhline{0.6pt}
        \end{tabular}
    \end{minipage}
    \hfill
    \begin{minipage}[tb]{0.48\columnwidth}
        \centering
        \caption{Evaluation of 3D detection methods, C for camera and L for LiDAR}
        \label{tab:detection_3d}
        \footnotesize
        \begin{tabular}{c|c|ccc}
        \Xhline{0.6pt}
        \multirow{2}{*}{Modality} & \multirow{2}{*}{Method} & \multicolumn{3}{c}{AP$_{3D|0.5}$}  \\ &     & Veh.    & Cyc.    & Ped.  \\ \Xhline{0.6pt}
        C          & ImVoxelNet\cite{rukhovich2022imvoxelnet}  &53.5   & 24.6   & 18.3 \\
        C          & BEVHeight\cite{yang2023bevheight}  & 73.5   & 49.3   & 34.7 \\
        C          & MonoFlex\cite{zhang2021monoflex}   & 74.2   & 16.2   & 13.3 \\ \hline
        L          & PointPillars\cite{pointpillars}    & 75.8   & 64.1   & 54.2   \\ \hline
        C + L     & BEVFusion\cite{liu2023bevfusion}    & 80.1   & 65.9   & 63.8   \\
        \Xhline{0.6pt}
        \end{tabular}
    \end{minipage}    
\end{table}

\begin{table}[tb]
    \centering
    \begin{minipage}[b]{0.48\columnwidth}
        \centering
        \caption{Tracking metrics for AB3D with PointPillars\cite{pointpillars} as the detector.}
        \label{tab:tracking}
        \setlength{\tabcolsep}{3mm}
        \begin{tabular}{c|c|c}
            \Xhline{0.6pt}
            Method                 &  MOTA  &  MOTP \\ \hline
            AB3D\cite{ab3d}        & 0.443  &  0.615  \\ \Xhline{0.6pt}
        \end{tabular}
    \end{minipage}
    \hfill
    \begin{minipage}[b]{0.48\columnwidth}
        \centering
        \caption{Metrics for motion segmentation baseline, F for F-measure.}
        \label{tab:motion_seg}
        \begin{tabular}{c|c|c|c|c|c}
            \Xhline{0.6pt}
            Method   & IoU  & P     & R     & F & FPS\\ \hline
            RiWNet+ & 0.691 & 0.741 & 0.717 & 0.726 & 11.3 \\ \Xhline{0.6pt}
        \end{tabular}
    \end{minipage}    
\end{table}

We then performed experiments on 3D detection methods for each modality. Although BEVFusion outperformed PointPillars in terms of the average precision score as shown in Table \ref{tab:detection_3d}, it is constrained to a limited detection range, and only applicable to roadside sites having both sensor modalities attached, not to mention the computation cost. PointPillars however, With a customized detection range and voxel size, is able to reach an accuracy for vehicles rival to BEVFusion and higher scores compared to those reported in \cite{yu2022dairv2x}. Interestingly, despite the variations in the number of beams, heights, and pose angles of the LiDARs concerning local ground in the CORP dataset, it has been observed that the spatial alignment of the point clouds to their LiDAR-base coordinate systems facilitates the accuracy and generalization of the model, we conjecture that the LiDAR-base coordinate system helps mitigate the differences in positioning and pose of LiDARs relative to the ground. Demonstrations of this assumption are left to researchers for further exploration.

\begin{figure}[t]
\centering
\includegraphics[width=1.0\columnwidth]{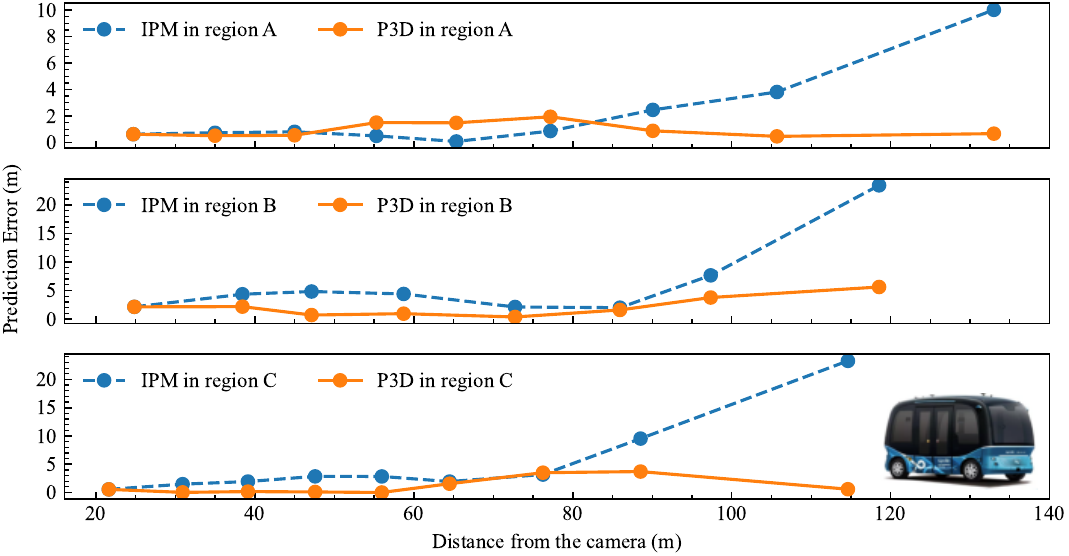}
\caption{A performance comparison beween P3D and IPM. The $x$ axis is the true distance from a target vehicle (shown at the bottom right) to the camera, while the $y$ axis is the absolute distance between the predicted and the ground-truth target-camera distance, with P3D results colored in black and IPM in red. Results are shown in \textit{Easy}, \textit{Medimu}, and \textit{Hard} scenarios individually from top to bottom.}
\label{fig:p3d_vs_ipm}
\end{figure}

\subsubsection{Object tracking.} Tracking with AB3D based on detection results from PointPillars is experimented as shown in Table \ref{tab:tracking}. Combined with Table \ref{tab:detection_3d}, one can expect better tracking scores with improved detection performance.\\

\subsubsection{Moving Object Segmentation.} We explore to adopt pixel segmentation tasks for moving targets in a campus setting. As shown in Table \ref{tab:motion_seg}, our RiWNet+ method achieved a baseline performance with a real-time rate of 11 fps. \cref{fig:motion_seg} illustrates several demanding scenarios for this task, where (a) is a scene showing the stark contrast between light and dark, (b) presents a case in which stationary targets dominate, (c) is a picture taken in the evening and (d) is from a camera with infrared mode turned on. A clear-cut and accurate pixel segmentation for moving targets remains ambitious in such situations.

\subsubsection{Distance Estimation.} For the evaluation of P3D, we compared it with the traditional yet industrially-recognized IPM method\cite{massimo1998IPM} in terms of prediction error in the distance from the camera to the target. For a fair comparison, images from 18 cameras in CORP are examined within their corresponding region, categorized into A, B and C based on the detection range. An experiment vehicle equipped with RTK GPS is driven through the FOV of each camera slowly, then the distance from the vehicle to the camera is predicted by P3D and IPM independently. As shown in \cref{fig:p3d_vs_ipm}, although for region A, the prediction error of P3D is larger in the middle but the overall error is less than 2 meters for that interval. Also we observed that P3D consistently outperformed IPM across all scenarios featuring diverse ground surfaces, camera height and pose angles, thereby establishing itself as a robust baseline for object-level distance estimation tasks. Again, here we focus on non-learning distance estimation methods.

\section{Conclusion}\label{sec:conclusion}
In this work, we present CORP, the first publicly available testbed designed for campus-oriented roadside perception tasks with a volume of 205k+ images and 102k+ point clouds captured from diverse camera and LiDAR sensors. The dataset is annotated by 2D and 3D bounding boxes with the unique ID assigned to the same targets captured by different sensors, and pixel-wise segmentation masks for moving targets. CORP extends the multi-modal detection of objects with the support for distant persistent tracking and pixel segmentation of moving targets. Moreover, we evaluated several deep-learning baseline perception algorithms to identify challenges in a campus setting and proposed a learning-free baseline for per-object distance estimation. Unlike other roadside datasets focusing on urban traffic, CORP 
broadens the scope to underscore the challenges for perception within campus and residential environments.



\bibliographystyle{unsrt}
\bibliography{references}

\clearpage

\appendix
\section*{Appendix}

\subsection{Rationale for P3D}\label{sec:about_p3d}
\begin{figure}[hb]
  \centering
    \includegraphics[width=0.8\columnwidth]{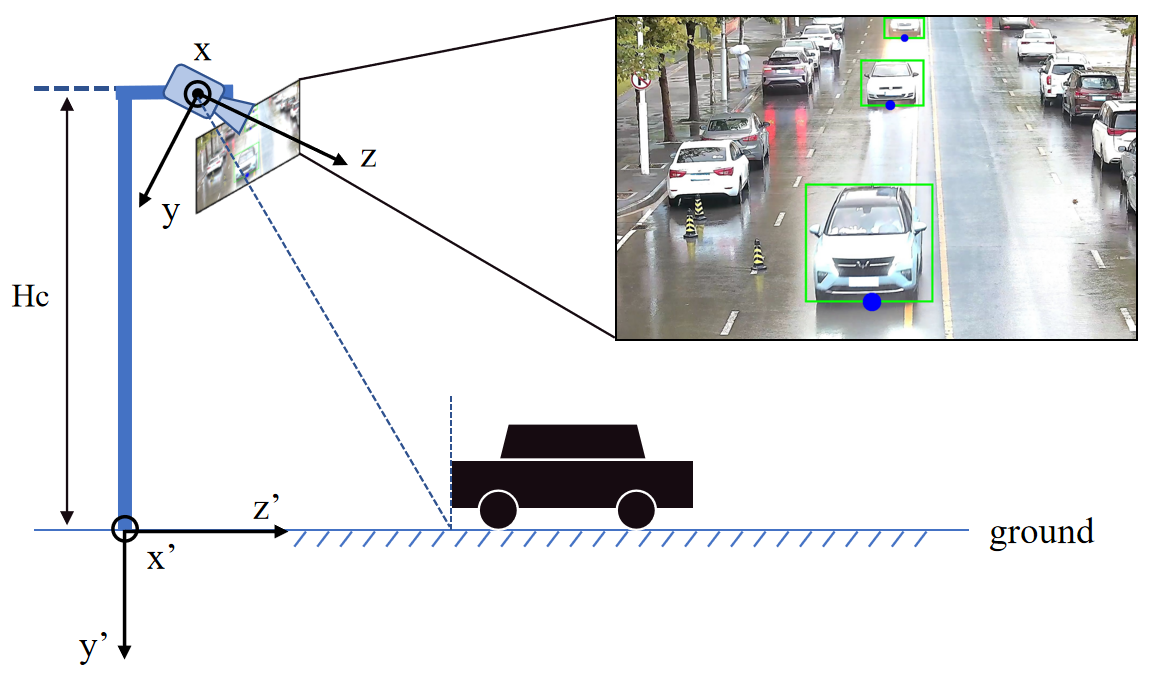}
  \caption{An illustration of camera and roadside coordinate systems. The camera coordinate system is denoted as $o$-$x$-$y$-$z$, which originates from the center of the camera with $Z$-axis pointing along the view and $x$-axis out of but not necessarily perpendicular to the screen. The road coordinate system is denoted as $O'$-$x'$-$y'$-$z'$ with $O'$ being the contact point between the pole and the ground, $z'$-axis is in the ground plane and in the same perpendicular plane with $z$-axis, while $y'$-axis is normal to the local ground surface. The green bounding boxes in the zoomed-in image are examples of 2D detection results, and the blue points on the lower boundaries of the boxes represent the car-road contact points, which are used in P3D for depth estimation.}
  \label{fig:coordinate_systems}
\end{figure}

In roadside scenarios, the camera sensor is usually positioned at a certain height $H_{c}$ above a local ground plane. We define the camera and the road coordinate systems as shown in Figure \cref{fig:coordinate_systems}. By employing P3D, the objective we aspire to achieve is to lift the pertinent 2D objects into real-world environment by the reconstruction of their 3D coordinates, in a non-learning fashion.

Admittedly, compared with end-to-end monocular 3D detection methods, the physical sizes and the orientation of targets are missing in P3D which employed 2D detectors. However, advantages of P3D over monocular 3D detectors are non-trivial: (i) it is less demanding for labeled data. (ii) 2D detector arguably have better domain adaptation ability than their 3D counterparts. (iii) 2D detectors require less computational resources and therefore are more efficient for large-scale implementation.

\subsection{Methods}
In this section, we first introduce the basic principle of the pinhole camera model with explicit assumptions and conclusions, then we describe the implementation details of P3D further.

\subsubsection{Geometry-based Depth Estimation Model}\label{sec:camera_model}

The pinhole camera model establishes the relationship between the road coordinate system and the pixel plane, by utilizing the presumably known intrinsic parameters obtained from the calibration process. We denote the camera intrinsic parameter using K as follows:

\begin{equation}\label{eq:K}
K = 
\begin{bmatrix}
f_{u} & 0 & c_{u} \\
0 & f_{v} & c_{v} \\
0 & 0 & 1
\end{bmatrix}
\end{equation}

To describe the ground plane in the camera's frame, we use the transformation matrix $T_{road2cam}\in R^{4\times 4}$ to establish the relationship between the road and the camera coordinates. Any 3D point in the road system $[x, y, z]_{road}^T$ can be transformed to the image plane to obtain their pixel coordinates $[u, v]^T$ via the following two steps:

\begin{equation}\label{eq:road2cam}
\begin{bmatrix}
   x\\
   y\\
   z\\
   1
\end{bmatrix}_{cam}=
T_{cam2road}^{-1}\cdot
\begin{bmatrix}
    x\\
    y\\
    z\\
    1\\
\end{bmatrix}_{road}
\end{equation}

\begin{equation}\label{eq:xyz2uv}
\begin{bmatrix}
   u\\
   v\\
   1
\end{bmatrix}
= K\cdot
\begin{bmatrix}
    x/z\\
    y/z\\
    1\\
\end{bmatrix}_{cam}
\end{equation}
here we already applied $T_{road2cam}=T_{cam2road}^{-1}$. 

The transformation matrix from the coordinate system to the road coordinate system can be decomposed as follows.
\begin{equation}
    T_{cam2road}=
    \begin{bmatrix}
    R & t\\
    0 & 1
    \end{bmatrix}
\end{equation}

Where $R$ and $t$ represent the rotational and translational transformation of a point from the camera coordinate system to the road system. The rotation matrix $R$ can be decomposed using the pitch and roll angles, denoted as $\alpha$ and $\gamma$ respectively, of the camera relative to the roadside reference frame:
\begin{equation}
    R = \begin{bmatrix}
        \cos{\gamma} & -\sin{\gamma} & 0\\
        \sin{\gamma} & \cos{\gamma} & 0\\
        0 & 0 & 1
    \end{bmatrix}\cdot
    \begin{bmatrix}
        1 & 0 & 0\\
        0 & \cos{\alpha} & -\sin{\alpha} \\
        0 & \sin{\alpha} & \cos{\alpha}
    \end{bmatrix}
    \equiv
    \begin{bmatrix}
        r_{ij}
    \end{bmatrix}_{3\times 3}
\end{equation}

In addition to the pinhole camera assumption, a second hypothesis similar to \cite{rope3d_2022_CVPR} for P3D is that the road surface, which is not necessarily flat, can be approximated as a group of planes along $z'$-axis in the road coordinate system, therefore they are also planar surfaces in the camera system $o$-$x$-$y$-$z$ with different pitch and roll angles w.r.t. the $x$-$o$-$z$ plane.

Consider a local plane defined by $o'$-$x'$-$z'$, the intersection line between the ground plane and the $y$-$o$-$z$ plane of the camera coordinate system serves as the $z$-axis of the road coordinate system, by which the influence of the camera's yaw angle in the road coordinates is eliminated and therefore not present. The pitch angle $\alpha$ indicates the angle of the camera's $z$-axis relative to the $z'$-axis of the road system. Similarly, the roll angle $\gamma$ represents the angle between $x'$ and $x$ axes. Then we we have:
\begin{equation}
    n^{T}\cdot Q_{cam} + d = 0
\end{equation}
where $Q_{cam}$ is the coordinates of a point in the camera system, $n = \begin{bmatrix}
    r_{21} & r_{22} & r_{23}
\end{bmatrix}^{T}$ is the normal vector to the plane and $d$ is the intercept parameter, which is $-H_{c}$ in the considered case. When it comes to a different plane, $d$, $\alpha$ and $\gamma$ should all be accounted for from that plane. By solving the above equations for $z$ in the case where $\gamma$ is close to 0, this depth $z$ can be approximated as:
\begin{equation}\label{eq:zpred_approx}
    z = \frac{H_{c}\cdot f_{v}}{-f_{v}\sin{\alpha} + (v-c_{v})\cos{\alpha}}
\end{equation}

Once $z$ of a target in an image is obtained with its pixel coordinates $[u, v]^T$, we can further get its Cartesian coordinates $[x, y, z]_{cam}^T$ in the camera system by the following: 

\begin{equation}\label{eq:z2xy}
\begin{bmatrix}
   x\\
   y\\
\end{bmatrix}
= 
\begin{bmatrix}
z/f_{u} & 0 & -z/f_{u}\cdot c_{u}\\
0 & z/f_{v} & -z/f_{v}\cdot c_{v}
\end{bmatrix}_{cam}
\cdot
\begin{bmatrix}
    u\\
    v\\
    1\\
\end{bmatrix}
\end{equation}

Worth mentioning, the vanishing point of a camera, defined by the intersection of two mutually parallel lines in $o$-$x$-$y$-$z$ coordinate system, is considered to have infinite depth, \ie the dominator in \cref{eq:zpred_approx} on the right side is zero, therefore we have:

\begin{equation}\label{eq:alpha}
    \tan{\alpha} = \frac{v_{vanish} - c_{v}}{f_{v}}
\end{equation}

Note, the above derivation is based on the assumption that the ground plane can be approximated by one or more geometrical planar surfaces, therefore the disparity between real-world scenarios in images and theoretical assumption serves as a catalyst for our classification of cameras into varying levels of difficulty for P3D.

\begin{figure}[htb]
  \centering
    \includegraphics[width=0.6\columnwidth]{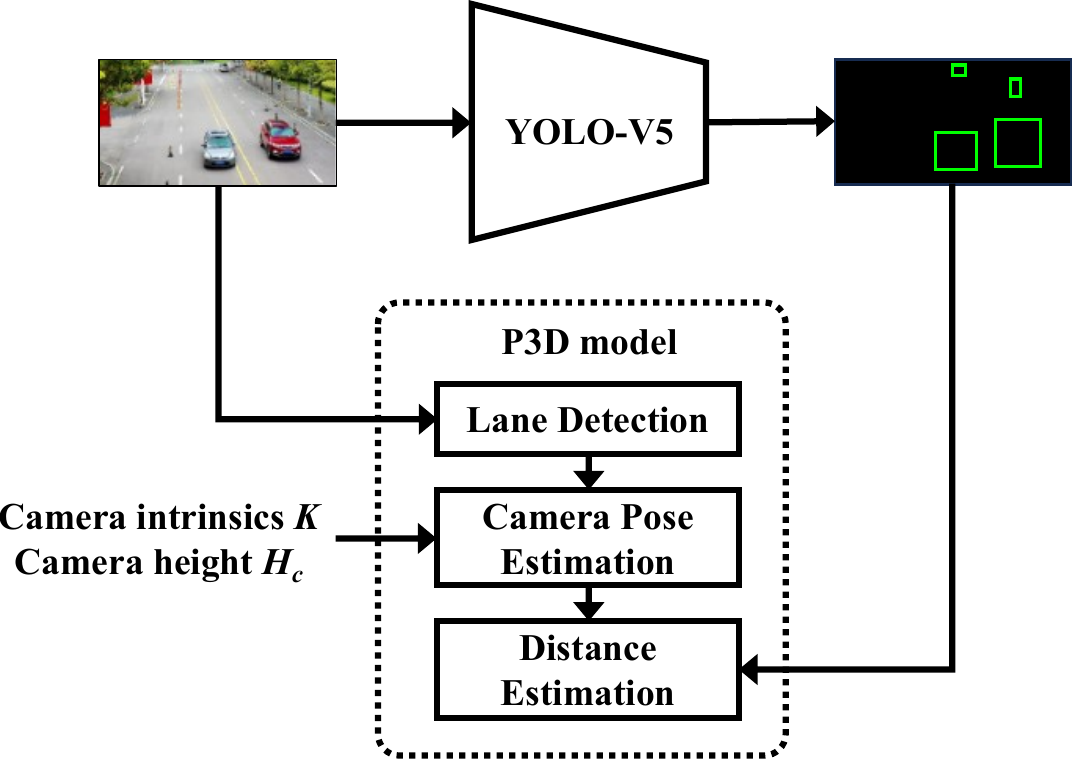}
  \caption{Overview of our P3D method.}
  \label{fig:p3d_overview}
\end{figure}

For the completeness of description, the overview of our Pseudo-3D (P3D) method for target distance estimation is shown in \cref{fig:p3d_overview}. The processing pipeline of P3D is the following: (i) with a given image, perform line feature detection for parallel lines, which in most cases are lane markings, then find their intersections as the vanishing point. (ii) using the vanishing point to obtain the pitch angle of a camera based on \cref{eq:alpha}. (iii) with the given pixel coordinates $[u, v]^T$ of a target, compute the estimated depth $z$ using \cref{eq:zpred_approx} and $x$, $y$ using \cref{eq:z2xy}.

\subsubsection{Implementation details}

Once the intrinsic parameters, pose angles and the height of camera sensors are measured, we can lift a 2D target in an image to a 3D point in the camera coordinate system by following the closed-form \cref{eq:zpred_approx} and \cref{eq:z2xy} with no computational cost, given an image-based 2D detector employed beforehand to produce the bounding boxes of interested targets. Subsequently, the midpoint of the lower edge of a box is selected as the contact point between the target and the ground, of which the pixel coordinates are used to calculate target depth. It is important to note that an accurate measurement of the height and the pitch angle for the camera are necessary to the P3D method. The overall algorithm is described in \cref{alg:p3d_psedo_code}.

\begin{algorithm}[htb]
    \begin{algorithmic}[1]
        \renewcommand{\algorithmicrequire}{\textbf{Input:}}
        \REQUIRE $ $\\
        $I$: current image frame;\\
        $K \in R^{3\times3}$: intrinsic matrix of the camera;\\
        $H_{C}\in R$: height of the camera w.r.t. the ground;\\
        $\alpha$: pitch angle of the camera w.r.t. the ground;\\
        $\gamma$: roll angle of the camera w.r.t. the ground;\\
        $T_{N}^{2D}\in R^{N\times2}$: the detection results from any 2D detector applied to the image, \ie  \ pixel coordinates of the target-road contact points.
        \renewcommand{\algorithmicrequire}{\textbf{Output:}}
	\REQUIRE $ $\\
        $\hat{T}_{N}^{3D} \in R^{N\times3}$: estimated 3D detection results.
        \renewcommand{\algorithmicrequire}{\textbf{Begin:}}
        \REQUIRE
        \STATE $\hat{T}_{N}^{3D} = \{ \}$
        \FOR{$t \in R^{1\times 2}$ in $T_{N}^{2D}$}
            \STATE $u, v \leftarrow t$
            \IF{$u, v$ beyond the boundaries of $I$}
                \STATE continue
            \ENDIF
            \STATE $z \leftarrow f(K, H_{C}, \alpha, \gamma)$ using \cref{eq:zpred_approx}
            \STATE $x \leftarrow (u-C_{u})\times z / f_{u}$ using \cref{eq:z2xy}
            \STATE $y \leftarrow (v-C_{v})\times z / f_{v}$ using \cref{eq:z2xy}
            \STATE $\hat{T}_{N}^{3D} \leftarrow \hat{T}_{N}^{3D} \cup \{x, y, z\}$
        \ENDFOR
        \RETURN $\hat{T}^{3D}$
    \end{algorithmic}
    \caption{P3D$(I, K, H_{C}, \alpha, \gamma, T_{N}^{2D})$}
    \label{alg:p3d_psedo_code}
\end{algorithm}

\subsubsection{Experimental Setup}

\begin{figure}[ht]
\centering
\begin{minipage}[b]{1.0\columnwidth}
\centering
\subfloat[][Change in altitude w.r.t. longitudinal displacement]{\includegraphics[width=1.0\columnwidth]{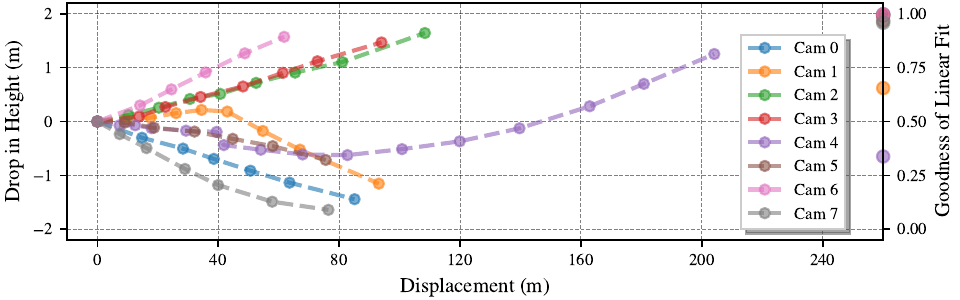}}
\vspace{5mm}
\end{minipage}\\

\begin{minipage}[b]{0.45\columnwidth}
\centering
\subfloat[][Even]{\includegraphics[width=1.0\columnwidth]{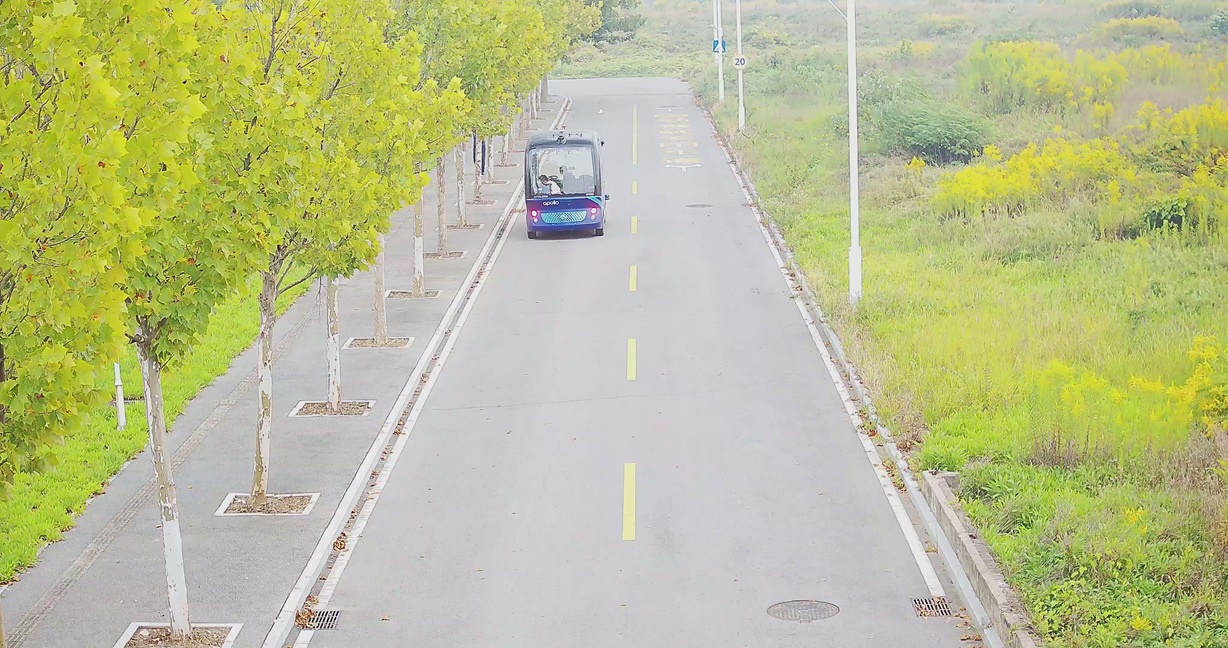}}
\end{minipage}
\begin{minipage}[b]{0.45\columnwidth}
\centering
\subfloat[][IPM counterpart of the 'even' image]{\includegraphics[width=1.0\columnwidth]{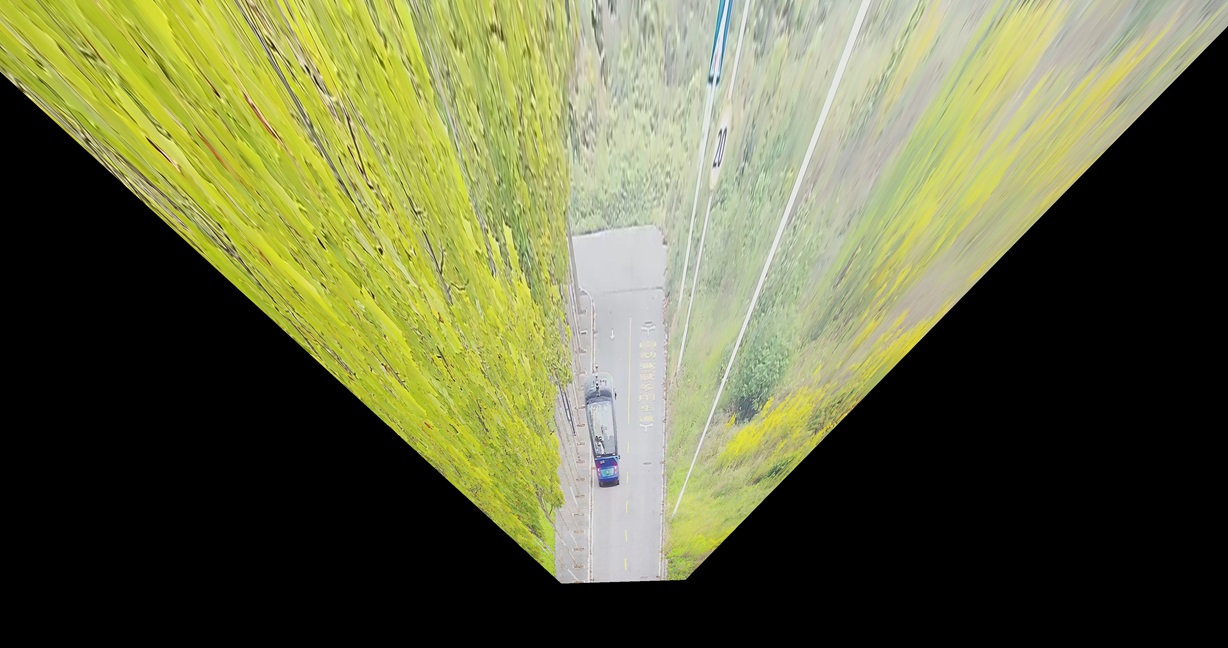}}
\end{minipage}\\

\begin{minipage}[b]{0.45\columnwidth}
\centering
\subfloat[][Partially-even]{\includegraphics[width=1.0\columnwidth]{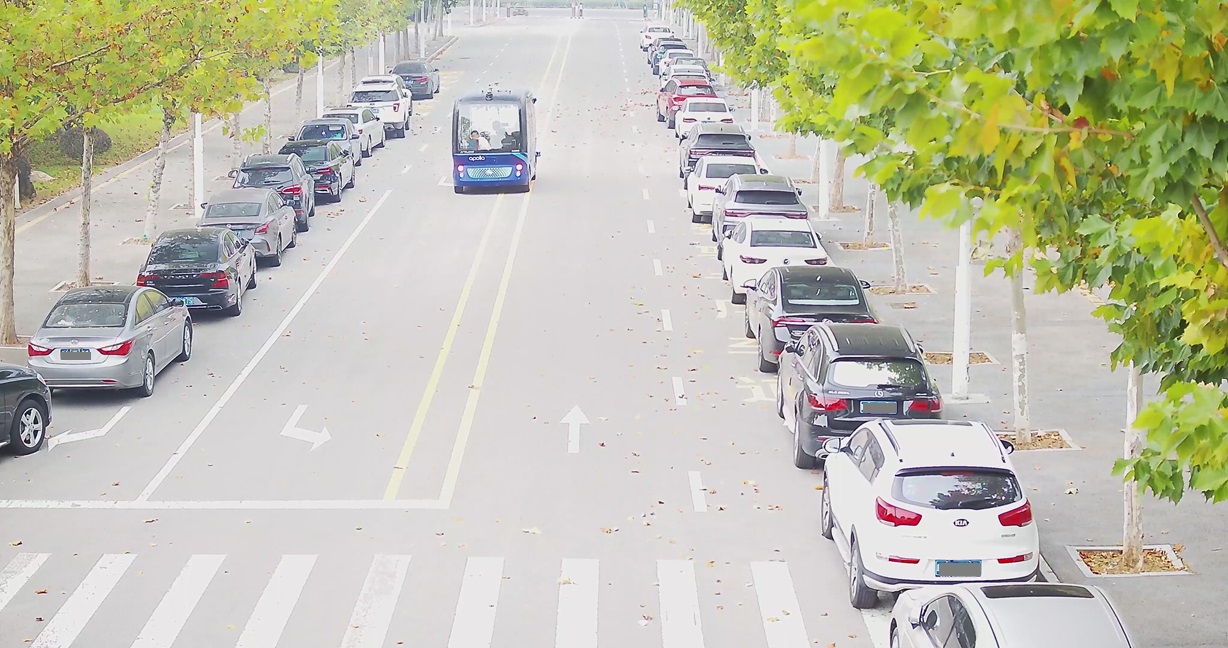}}
\end{minipage}
\begin{minipage}[b]{0.45\columnwidth}
\subfloat[][IPM counterpart of the 'partially-even' sample]{\includegraphics[width=1.0\columnwidth]{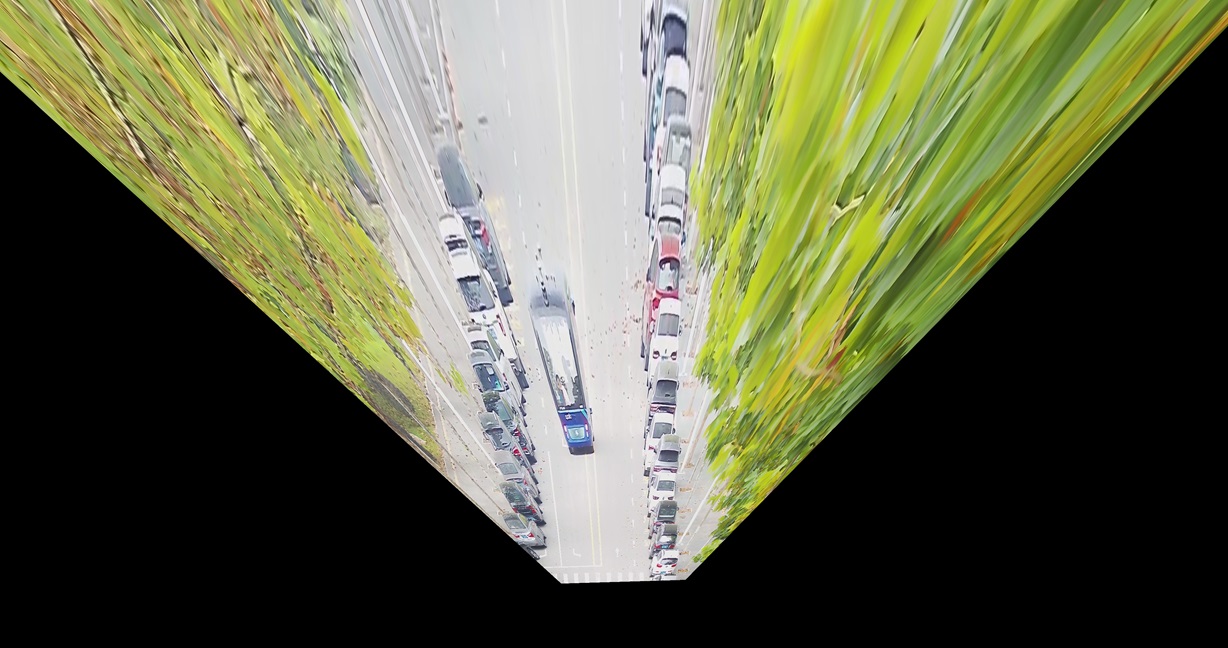}}
\end{minipage}\\

\begin{minipage}[b]{0.45\columnwidth}
\centering
\subfloat[][Uneven]{\includegraphics[width=1.0\columnwidth]{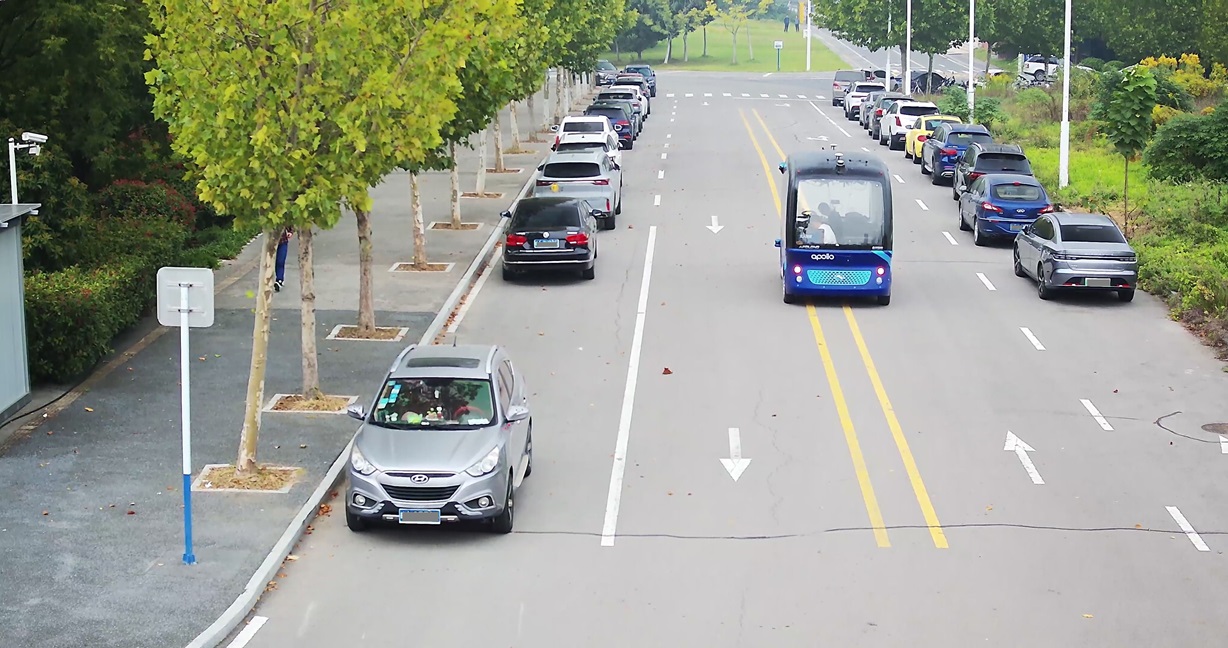}}
\end{minipage}
\begin{minipage}[b]{0.45\columnwidth}
\subfloat[][IPM counterpart of the 'uneven' image]{\includegraphics[width=1.0\columnwidth]{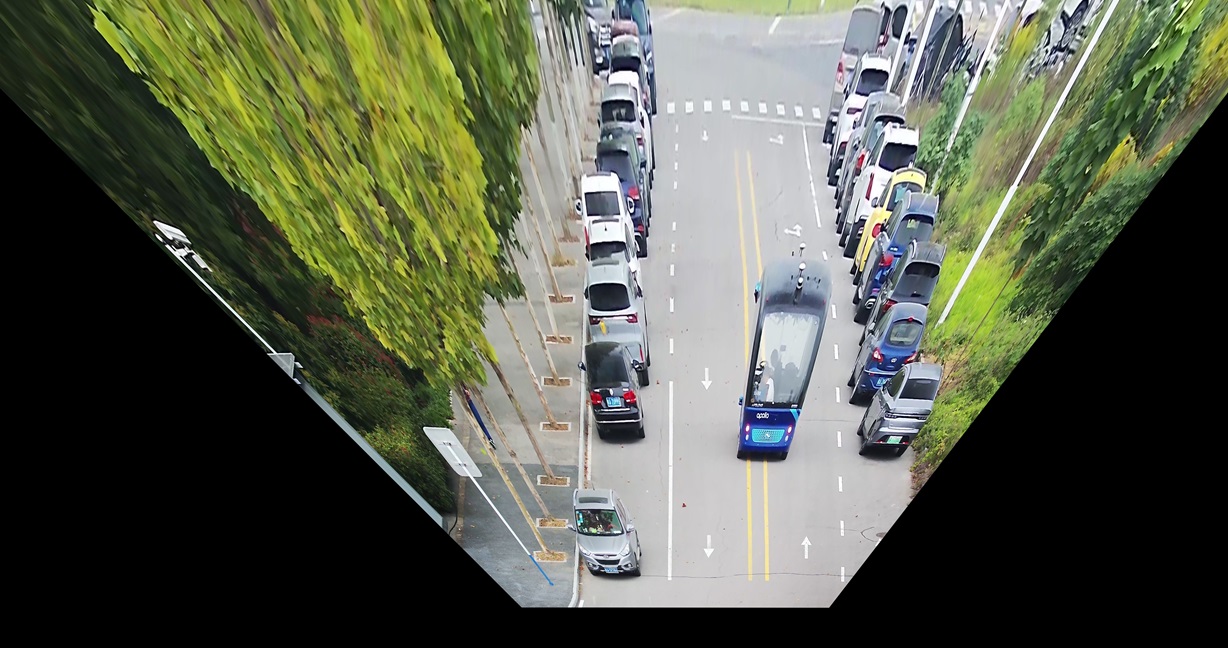}}
\end{minipage}

\caption{Flatness profiles of the typical ground surfaces in the dataset and sample images for cameras in the \textit{Easy}(b,c), \textit{Medium}(d,e) \textit{Hard}(f,g) categories.}
\label{fig:easy_medium_hard}
\end{figure}

For a through comparison between P3D and IPM methods, we sampled the camera images from all of the 3 regions mentioned in the main text, \ie \textit{A}, \textit{B} and \textit{C}. In addition, we took a closer look at the flatness of the ground surface in the view of individual cameras, which is measured by a vehicle equipped with RTK GPS device. Specifically in each trial, we approached the camera by driving a minibus, continuously decreasing its distance from afar until it reached close proximity. We recorded the longitudinal trajectory of the vehicle and performed analysis of its displacements along the course.

As shown in \cref{fig:easy_medium_hard}(a), we depict the change in the $z$-component of the car's coordinates against its longitudinal displacements for each experiment, with the score for the goodness of linear fit shown on the right side in (a), as an indicator of the ground flatness. Note that the detection range varies across the sampled cameras due to environmental occlusions and road corners, and the goodness of fit scores can be easily clustered into \textit{even}, \textit{partially-even} and \textit{uneven}. Sample images of the experimental scenes can be found in \cref{fig:easy_medium_hard}(b)~(g).

\end{document}